\documentclass{article}
\usepackage{natbib}
\usepackage{microtype}
\usepackage{graphicx}
\usepackage{subfigure}
\usepackage{booktabs} 
\usepackage{amsmath, amssymb, amsthm, mathtools}
\usepackage{hyperref}
\usepackage[capitalize,noabbrev]{cleveref}
\usepackage{multirow}
\usepackage{balance}
\usepackage{enumitem}
\usepackage{listings}
\usepackage{stfloats} 
\usepackage[colorinlistoftodos]{todonotes}
\usepackage[titletoc]{appendix}
\usepackage[accepted]{icml2024}

\begin{document}

\twocolumn[
\icmltitle{Reasoning-Based AI for Startup Evaluation (R.A.I.S.E.): A Memory-Augmented, Multi-Step Decision Framework}

\begin{icmlauthorlist}
\icmlauthor{Jack Preuveneers\texorpdfstring{\textsuperscript{(1)}}{(1)}}{}
\icmlauthor{Joseph Ternasky\texorpdfstring{\textsuperscript{(2)}}{(2)}}{}
\icmlauthor{Fuat Alican\texorpdfstring{\textsuperscript{(2)}}{(2)}}{}
\icmlauthor{Yigit Ihlamur\texorpdfstring{\textsuperscript{(2)}}{(2)}}{}
\end{icmlauthorlist}

\begin{center}
\footnotesize
\texorpdfstring{\textsuperscript{(1)} University of Oxford}{(1) University of Oxford} \quad
\texorpdfstring{\textsuperscript{(2)} Vela Research}{(2) Vela Research}
\end{center}
\vskip 0.15in

\icmlcorrespondingauthor{Jack Preuveneers and Yigit Ihlamur}{jack.preuveneers@eng.ox.ac.uk and yigit@vela.partners}

\icmlkeywords{LLM, Decision Policy, Venture Capital, Explainability, Memory, Refinement, Reinforcement Learning}

\vskip 0.3in ]

\printAffiliationsAndNotice{ }

\begin{abstract}
We present a novel framework that bridges the gap between the interpretability of decision trees and the advanced reasoning capabilities of large language models (LLMs) to predict startup success. Our approach leverages chain-of-thought prompting to generate detailed reasoning logs, which are subsequently distilled into structured, human-understandable logical rules. The pipeline integrates multiple enhancements—efficient data ingestion, a two-step refinement process, ensemble candidate sampling, simulated reinforcement learning scoring, and persistent memory—to ensure both stable decision-making and transparent output. Experimental evaluations on curated startup datasets demonstrate that our combined pipeline improves precision by 54\% from 0.225 to 0.346 and accuracy by 50\% from 0.46 to 0.70 compared to a standalone OpenAI o3 model. Notably, our model achieves over 2× the precision of a random classifier (16\%). By combining state-of-the-art AI reasoning with explicit rule-based explanations, our method not only augments traditional decision-making processes but also facilitates expert intervention and continuous policy refinement. This work lays the foundation for the implementation of interpretable LLM-powered decision frameworks in high-stakes investment environments and other domains that require transparent and data-driven insights.
\end{abstract}

\section{Introduction}

Early-stage venture capital investment is a high-risk, high-reward domain where investors must identify potential ``unicorn'' startups with limited information. Traditional decision trees offer some interpretability but struggle with the complexity of non-linear data, while large language models (LLMs) provide powerful reasoning capabilities at the cost of transparency. Our project addresses these challenges by introducing a novel framework for startup founder evaluation that combines LLM-based reasoning with structured, rule-based decision-making. This integrated approach not only achieves high predictive accuracy but also produces clear, human-understandable explanations for each prediction.

Our primary objective is to build an explainable investment model that outperforms random selection by over 10× while remaining fully transparent and editable by experts. Unlike opaque black-box models, our framework allows venture capitalists to understand and, if necessary, override the model's decisions using their domain expertise. By transforming LLM-generated reasoning logs into explicit and verifiable rules, our system empowers decision-makers to review, adjust, and backtest the underlying policies, ensuring that each version can be refined over time. This level of adaptability should foster greater trust among stakeholders and enable a continuous feedback loop that ultimately leads to more resilient and effective investment strategies.

The framework is built on a modular and iterative design that integrates several advanced techniques. It incorporates multistep refinement to enhance the quality of the LLM's chain-of-thought reasoning, ensemble candidate sampling to reduce variability in predictions, simulated reinforcement learning (RL)-based scoring to further refine output quality, and persistent conversational memory to maintain context over multiple interactions. Each component contributes uniquely to improving overall performance, ensuring that every prediction is reliable and interpretable.

In this research, we demonstrate that LLMs, when guided to reason and explain their thought process, can serve as powerful allies in complex decision domains. Our code-based framework, implemented with the \texttt{o3-mini} model, systematically generates natural-language reasoning logs, extracts structured rules, and compiles these into an interpretable decision policy for startup success prediction. This work not only overcomes the rigidity and opacity of conventional machine learning models, but also lays the groundwork for future research and practical applications.

\section{Literature Review}
Our work builds on multiple strands of prior research. Recent advances in LLMs have opened new avenues for predictive analytics in domains such as venture capital (VC). However, the opaque nature of such models conflicts with the need for explainability in high-stakes investment decisions. Traditional approaches to predicting early-stage company success often rely on structured data and ensemble learning. While they achieve good accuracy, they lack interpretability. 

\cite{xiong2023founder} proposed a novel approach called Founder-GPT that measures how well a startup’s business concept aligns with its founder's unique profile. Instead of treating all ventures uniformly, the framework integrates advanced LLM techniques, including self-simulation, iterative tree-based reasoning, and subsequent evaluation, to capture the nuanced interplay between a founder's background and the idea. Initial experiments indicate that this tailored analysis provides valuable insights, suggesting that evaluating startups through a personalised and founder-based lens can markedly improve predictions of future success. \cite{Xiong2024} also addresses the interpretability–performance trade-off with \textsc{GPTree}, an LLM-powered decision tree framework. GPTree combines the explainability of decision trees with the reasoning ability of LLMs, using a task-specific prompt to drive tree splits. Each node’s decision criterion is generated by the LLM, yielding human-readable rules at each branch. The framework also integrates an \emph{expert-in-the-loop} mechanism for human refinement of the tree after initial training \cite{Xiong2024}. This hybrid approach significantly outperformed both few-shot GPT-4 and human investors at identifying “unicorn” startups, achieving 7.8\% precision for early-stage unicorn prediction (vs.\ 3–5\% for humans), while providing transparent decision logic. 

Other works enhance LLM transparency and accuracy via reasoning traces and self-refinement. \cite{Kashyap2024} use a two-stage prompting approach: the LLM first produces a step-by-step chain-of-thought analysis of the input, then refines its initial answer based on that analysis. This “Thought Refinement” technique boosts the model’s recall from 53\% to 75\% \cite{Kashyap2024}. Similarly \cite{Muennighoff2025} propose a test-time scaling method (“s1”) that prevents the model from terminating its reasoning prematurely. By prompting the model to “wait” and extend its chain-of-thought, the LLM can double-check and correct errors, yielding up to 27\% accuracy gains on complex questions \cite{Muennighoff2025}. Although not specific to VC, such explicit reasoning steps and iterative self-correction could make investment predictions more interpretable and reliable. 

Ensemble methods offer another route to reliability. \cite{Schoenegger2024} demonstrate that an ensemble of 12 LLMs (a “silicon crowd”) can rival human judgement: their aggregated predictions on 31 forecasting tasks were as accurate as those of a crowd of 925 human forecasters, outperforming individual models and chance. These results suggest that ensembling multiple reasoning agents yields a strong predictor, analogous to the wisdom-of-crowds effect. \cite{Crescas2024} apply a random forest to classify startups as successful or not, achieving 91\% recall and 7\% higher accuracy than prior studies. Yet while ensembles improve performance through aggregation, their decision processes remain difficult to interpret, posing challenges for VC practitioners that require transparent rationales.

Another key trend is the fusion of LLMs with structured financial data and ensemble methods. \cite{maarouf2025fused} introduce a “fused” LLM that combines startup textual profiles with fundamental features (e.g., founding date, sector) to predict venture success. This hybrid model outperforms text-only baselines by a large margin, achieving an AUROC above 0.82 and higher return-on-investment forecasts. The contribution of textual self-descriptions was found to be substantial, raising predictive accuracy by about 2.2 percentage points when added to fundamentals. Similarly, \cite{ozince2024automating} leverages LLM-based prompting to engineer new features for startup evaluation. Their framework uses chain-of-thought prompts to segment founder attributes (e.g., experience level, persona) from minimal data, which are then used in a predictive model. This approach reveals interpretable founder-success patterns and improves prediction accuracy, illustrating how LLMs can enrich traditional venture data with qualitative insights. 

In the realm of decision support, Wang developed \textit{Startup Success Forecasting Framework (SSFF)}, an AI-native VC analyst agent \cite{wang2024ssff}. SSFF combines classic machine learning models with an LLM-driven “analyst” module and external data retrieval to mimic a venture capitalist’s due diligence process \cite{wang2024ssff}. The system can ingest minimal information (as little as a company name or founder profile) and autonomously produce an investment recommendation, thanks to its multistep reasoning pipeline that integrates real-time information and even a dedicated model for founder–idea fit. By decomposing the task into an explainable pipeline (prediction blocks, analysis blocks, and knowledge blocks), SSFF offers a more transparent alternative to end-to-end black-box predictors, while matching human-level analysis quality. Notably, these hybrid and ensemble-based designs address the “black box” challenge by providing intermediate rationales (e.g., decision tree splits, justified prompts, or feature attributions) that stakeholders can inspect.

In summary, emerging LLM-powered investment frameworks strive to pair predictive power with explainability. The reviewed approaches tackle this from different angles: GPTree provides transparent rules with human oversight; chain-of-thought prompting lets models explain and refine their reasoning; and ensemble strategies reduce variance via crowd consensus. While these methods show promise, they also have limitations. Future work may combine these techniques (e.g. multiple LLMs cross-verifying reasoning) and incorporate domain-specific knowledge to guide the AI.

Our framework integrates these ideas to produce an explainable high-precision model for the prediction of startup success.

\section{Methodology}
\begin{figure}[ht!]
    \centering
    \includegraphics[width=0.4\textwidth]{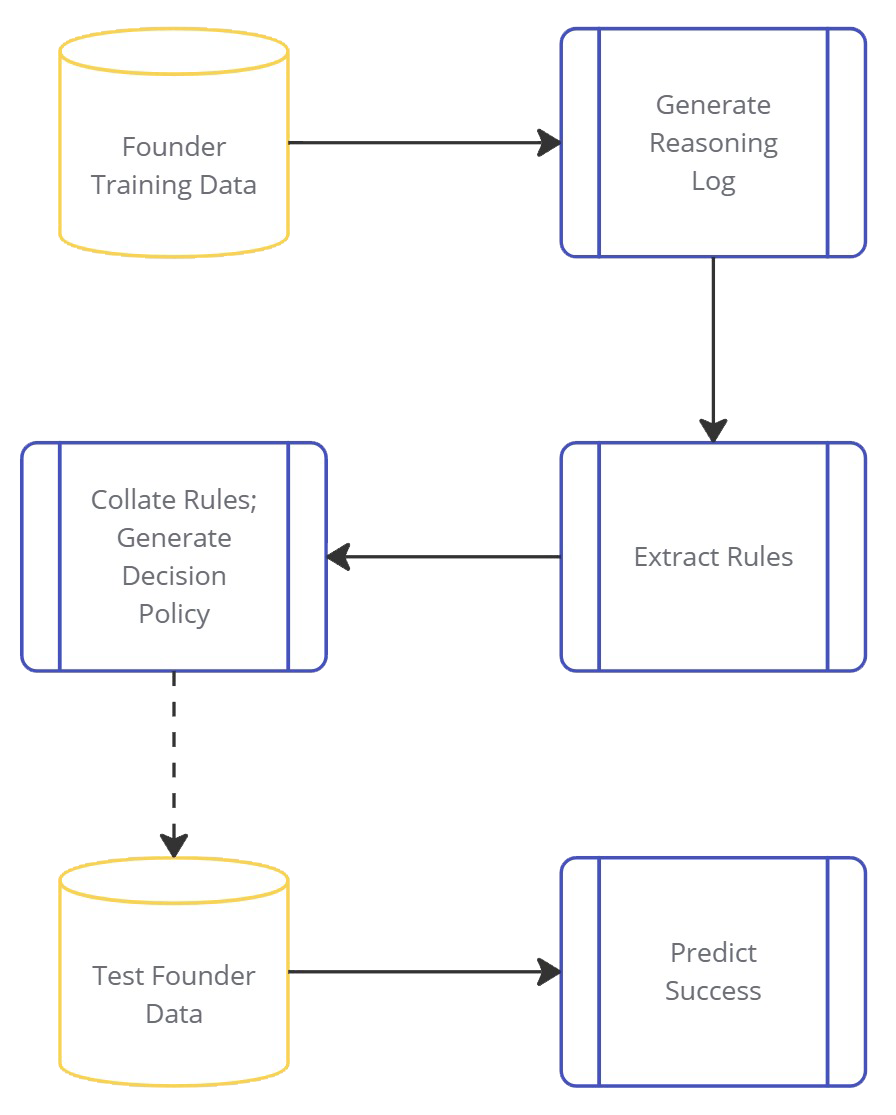}
    \caption{Overview of the LLM-Powered Investment Decision Framework Pipeline.}
    \label{fig:pipeline}
\end{figure}

Our base pipeline consists of three main stages, as illustrated in \Cref{fig:pipeline}. In the sections below, we detail each component, with additional improvement modules.

\subsection{Founder Training Data Ingestion and Preprocessing}
Efficient data handling is crucial given the size and heterogeneity of founder profiles. We implement a chunked reading approach using a load-and-preprocess function. This function does the following tasks:
\begin{enumerate}
    \item Reads the CSV file in manageable chunks to minimise memory overhead.
    \item Concatenates textual fields into a unified profile text:
    \begin{itemize}
        \item \texttt{clean\_linkedin\_profile}
        \item \texttt{clean\_cb\_profile}
        \item \texttt{company\_description}
    \end{itemize}
    \item Logs progress and aggregates the processed chunks into a single DataFrame.
\end{enumerate}

This approach ensures that even very large datasets can be handled efficiently without sacrificing data quality.

\subsection{LLM-Based Reasoning Log Generation}
A critical component of our system is generating detailed chain-of-thought reasoning logs for each founder. We begin by taking a task-specific string as user input to serve as context for the LLM. By concatenating the founder's LinkedIn profile, Crunchbase summary, and company description into a single prompt, the system clearly defines the context. This allows the LLM to approach the task as if it were an expert startup analyst, using its pre-trained knowledge to generate insightful reasoning.

For example, if our goal is to distinguish successful founders, the prompt might be as follows:

\texttt{"You are an expert startup analyst. Given a founder's background and startup description, provide a concise, clear, structured reflection explaining the key reason(s) for founder success or failure."}

The prompt instructs the LLM to provide a step-by-step chain-of-thought explanation for why the startup was successful or unsuccessful. This focused instruction helps the LLM prioritise the most relevant details from the founder's background and venture description, reducing extraneous output.

For each founder, we construct a prompt that includes the following items:
\begin{itemize}
    \item \textbf{Founder Profile:} Aggregated text from LinkedIn and Crunchbase.
    \item \textbf{Startup Description:} Detailed textual description of the venture.
    \item \textbf{Outcome Information:} A label indicating whether the startup was successful or unsuccessful.
    \item \textbf{Task Instruction:} A request for a step-by-step explanation of the factors contributing to the outcome.
\end{itemize}

This prompt is sent to the OpenAI model (\texttt{o3-mini}) via our client, and the response—a detailed reasoning trace—is recorded along with token usage. This step is crucial for later converting the LLM’s reasoning into logical rules.

This function iterates over all founders in the training set; the LLM response is parsed to extract the reasoning text. The structured, task-specific prompt ensures that similar founder profiles yield consistent reasoning logs. Consistency in these outputs is crucial, as it forms the basis for reliably extracting structured logical rules later in the pipeline. For each founder we also track cumulative token usage (for cost estimation), and periodically save the logs to a CSV file in case of crashes.

Example answers taken from the LLM are shown below in an anonymised manner.

\textbf{Sample Answer Success:}\\
\textit{"John Smith has a top-tier technical education from MIT and participated in programmes at Y Combinator and CFAR, underpinning his strong technical foundation. His roles at prestigious organisations (such as Google Brain and OpenAI for GPT-3 development) have provided him with deep insights into cutting-edge AI research and engineering practices. His blend of technical excellence, entrepreneurial track record, and strategic networking made him particularly well-equipped to steer a startup like Anthropic towards success in the competitive AI landscape."}

\textbf{Sample Answer Failure:}\\
\textit{"Jane Doe's background is heavily rooted in healthcare and biotechnology research, particularly in wound care, hyperbaric medicine, and hospital sales. This experience provided a strong foundation in patient care and pharmaceutical/medical contexts rather than in agricultural or industrial production.
Although her MBA has a sustainability focus, her career history lacks direct experience in indoor farming operations, ag-tech engineering, or agribusiness, which are crucial for developing and scaling a novel insulated building for crop production."}

\subsection{Rule Extraction from Reasoning Logs}
To keep the LLM on track, we again begin by taking a task-specific string as user input to serve as context for the LLM:

\texttt{"Convert the reasoning log into a single concise logical rule about the founder, and founder only, using the specified format."}

Each reasoning log is processed to extract a structured, logical rule in the form: \newline\newline
\texttt{\detokenize{IF <conditions> THEN likelihood_of_success = <result>}}

A conversion prompt is constructed, including an example rule and the founder’s reasoning log. This prompt guides the model to produce such a rule. 

\textbf{Example Prompt:}
\noindent\texttt{Convert the following reasoning log into a structured logical rule explaining why the founder SUCCEEDED or FAILED using the format:}\\ 
\texttt{IF <conditions> THEN likelihood\_of\_success = <result>.}\\ 

\texttt{Example:}\\ 
\texttt{IF founder has a top-tier university background AND previous experience at a successful startup AND startup idea targets a rapidly growing industry THEN likelihood\_of\_success = HIGH.}\\ 
\texttt{OR}\\ 
\texttt{IF founder has no documented professional experience AND no previous entrepreneurial ventures AND lacks relevant industry knowledge THEN likelihood\_of\_success = LOW.}

If the LLM output is unsatisfactory, a fallback regex-based method is used to ensure that the rule accurately reflects the actual outcome.

The resulting rule, along with the actual outcome (stored in a separate column), is saved to a CSV file.

Example answers of the extracted rules taken from the LLM on the previously shown reasoning logs are shown below.

\textbf{Sample Rule Success:} 
\textit{``IF founder attended top-tier institutions (e.g., MIT, Harvard) AND held leadership roles in major tech companies (e.g., CTO at Stripe) AND maintained an extensive, high-profile network within the tech community AND actively engaged in the tech and AI ecosystem THEN likelihood\_of\_success = HIGH.''}

\textbf{Sample Rule Failure:} 
\textit{``IF founder's background is primarily in healthcare and biotechnology research AND lacks direct experience in indoor farming, ag-tech engineering, and large-scale construction operations required for a super-insulated building concept THEN likelihood\_of\_success = LOW.''}

\subsection{Decision Policy Generation}
Once the initial rules are compiled into a preliminary decision policy, the rules are separated into two groups: success and failure. Two separate prompts (one per group) are then sent to the LLM to generate a concise decision policy for each group. The resulting policies are subsequently combined into a unified policy document.

We provide the following task-specific context:
\noindent\texttt{Analyse the following extracted rules from successful/unsuccessful founder profiles and compile a concise decision policy that clearly summarises the key conditions which predict startup success/failure.}\\ 
\texttt{Rules: ...}\\ 
\texttt{Your output should be in the following format:}\\ 
\texttt{IF <conditions> THEN likelihood\_of\_success = HIGH/LOW.}

\subsection{Start Up Success Prediction}
The decision policy is then used to generate predictions on the test set. For each founder, the prediction function constructs a prompt that includes the founder’s profile and the decision policy. The LLM is queried to return a prediction (HIGH or LOW) along with an explanation. The final prediction, together with its explanation, is appended to the DataFrame.

To maintain focus, we again use a task-specific context:

\texttt{"Predict the likelihood of success (HIGH or LOW) based on the provided founder profile and decision policy, and provide a brief explanation of your reasoning."}

An example prompt is constructed as follows: \newline
\noindent\texttt{Founder Profile: [clean\_linkedin\_profile] | [clean\_cb\_profile]}\\ 
\texttt{Startup Description: [company\_description]}\\ 
\texttt{Based on the following decision policy, predict whether the founder is likely to succeed.}\\ 
\texttt{Decision Policy: <policy\_text>}\\ 
\texttt{Return your prediction in the following format:}\\ 
\texttt{Prediction: <HIGH or LOW> }\\ 
\texttt{Explanation: <brief explanation>}

\subsection{Evaluation and Metrics}

The framework’s predictions are compared with the actual outcomes using standard evaluation metrics:
\begin{itemize}
\item \textbf{Precision:} Precision is the ratio of correctly predicted positive observations to the total predicted positive observations. It is especially useful when the cost of false positives is high.
\[
\text{Precision} = \frac{TP}{TP + FP}
\]

\item \textbf{Recall (Sensitivity):} Recall is the ratio of correctly predicted positive observations to all actual positives. It is crucial when missing a positive instance (false negatives) is costly.
\[
\text{Recall} = \frac{TP}{TP + FN}
\]

\item \textbf{F1 Score:} The F1 Score is the harmonic mean of precision and recall. It provides a single measure that balances both precision and recall.
\[
\text{F1 Score} = 2 \times \frac{\text{Precision} \times \text{Recall}}{\text{Precision} + \text{Recall}}
\]

\item \textbf{Matthews Correlation Coefficient (MCC):} MCC is a balanced measure that takes into account true and false positives and negatives. It is regarded as a balanced metric even if the classes are of very different sizes. \newline

\resizebox{\linewidth}{!}{$
\text{MCC} =
\frac{TP \times TN - FP \times FN}{\sqrt{(TP+FP)(TP+FN)(TN+FP)(TN+FN)}}
$}

\item \textbf{Accuracy:} Accuracy measures the proportion of correctly classified samples among all tested samples.
\[
\text{Accuracy} = \frac{TP + TN}{TP + TN + FP + FN}
\]
\end{itemize}

Evaluation results, along with detailed prediction logs and RL scores, are saved to CSV files.

\section{Results}
We evaluated our startup prediction framework on a balanced dataset—100 success and 100 failure cases for training and a fixed test set of 10 successes and 50 failures—to ensure a fair comparison across different model enhancements. 
We evaluate a series of experiments that progressively enhance our startup prediction pipeline. Our methodology begins with a comparison between the older OpenAI \texttt{4o-mini} model and our current \texttt{o3-mini} model, followed by refinements that include a two-step reasoning process, ensemble candidate sampling (3 picks), simulated RL-based scoring, and the incorporation of persistent conversational memory. Each variation is assessed using standard metrics—precision, recall, F1 score, MCC, and overall accuracy—as well as confusion matrices to illustrate the model’s predictive distribution.

\subsection{4o vs o3}

In our initial experiments, we compared the performance of the older OpenAI \texttt{4o-mini} model against our current choice, the \texttt{o3-mini} model. The \texttt{4o-mini} model provided acceptable chain-of-thought responses; however, its outputs were generally more verbose and contained redundant or tangential information. In contrast, the \texttt{o3-mini} model demonstrated a sharper focus on the key elements of each founder’s profile and venture description. This resulted in more concise and coherent reasoning logs, which in turn made the subsequent rule extraction process more reliable. Our experiments showed that the \texttt{o3-mini} model not only improved token efficiency---thereby reducing API costs---but also yielded more interpretable outputs. These improvements were crucial for the framework, as the clarity of the reasoning log is directly linked to the quality of the structured rules extracted later.

\begin{table}[ht]
\centering
\begin{tabular}{lcc}
\toprule
\textbf{Metric} & \textbf{4o Model} & \textbf{o3-mini Model} \\
\midrule
Precision & 0.200 & 0.225 \\
Recall    & 0.900 & 0.900 \\
F1 Score  & 0.327 & 0.368 \\
MCC       & 0.155 & 0.221 \\
Accuracy  & 0.383 & 0.467 \\
\bottomrule
\end{tabular}
\caption{Performance Comparison: 4o vs. o3-mini}
\label{tab:4o_vs_o3}
\end{table}

To provide additional insight, we also compare confusion matrices for the \texttt{4o-mini} and \texttt{o3-mini} models. 
Table~\ref{tab:cm4o} and Table~\ref{tab:cmo3} illustrate how each model’s predictions are distributed across the ``Success'' and ``Failure'' categories on our test set of 60 founders (10 actual successes and 50 actual failures). Note that exact confusion-matrix numbers can sometimes be approximate or differ slightly from the reported global metrics (e.g., due to rounding or because the LLM can produce slightly different reasonings and rules on each run).

\begin{table}[ht]
\centering
\begin{tabular}{l|cc}
\multicolumn{1}{c}{} & \multicolumn{2}{c}{\textbf{Predicted}}\\
\cline{2-3}
 & Failure & Success \\
\hline
\textbf{Actual Failure} & 14 & 36 \\
\textbf{Actual Success} & 1 & 9 \\
\end{tabular}
\caption{Confusion Matrix for the 4o Model}
\label{tab:cm4o}
\end{table}

\begin{table}[ht]
\centering
\begin{tabular}{l|cc}
\multicolumn{1}{c}{} & \multicolumn{2}{c}{\textbf{Predicted}}\\
\cline{2-3}
 & Failure & Success \\
\hline
\textbf{Actual Failure} & 19 & 31 \\
\textbf{Actual Success} & 1 & 9 \\
\end{tabular}
\caption{Confusion Matrix for the o3-mini Model}
\label{tab:cmo3}
\end{table}

From these matrices, we see that the \texttt{o3-mini} model correctly identifies more negative cases (actual failures) than the \texttt{4o-mini} model and also maintains a slightly lower false-positive count on actual failures. This corresponds qualitatively with the higher accuracy and MCC scores reported in Table~\ref{tab:4o_vs_o3}.

\subsection{o3 vs two-step refinement }
\begin{figure}[ht!]
    \centering
    \includegraphics[width=1.07\columnwidth]{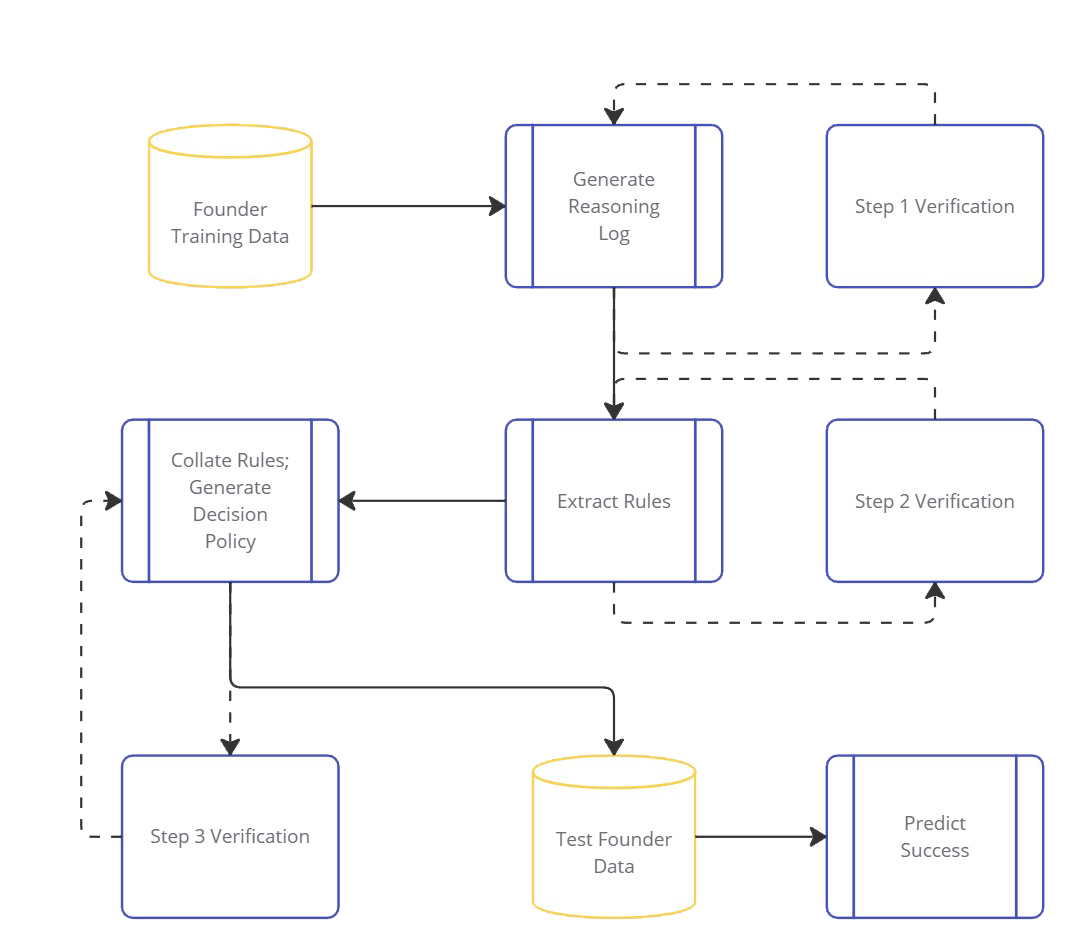}
    \caption{Overview of the Two-Step Reasoning Process.}
    \label{fig:2step}
\end{figure}

In this pipeline (shown in \Cref{fig:2step}) the \texttt{o3-mini} model is employed in a two-step refinement process. The first step involves generating a raw chain-of-thought reasoning log for each founder, while the second step refines this output to verify and clarify the key decision signals. This two-step approach ensures that only the most relevant and accurate details are retained for rule extraction. Compared to using a single direct pass with \texttt{o3-mini}, the two-step refinement substantially improves the consistency of the extracted rules and reduces noise in the reasoning logs. By iteratively checking what is correct, we ensure that only the most robust and pertinent conditions are retained, ultimately leading to more precise logical rules that better capture the underlying factors of founder success. 

An example prompt for this method is as follows:

\noindent\texttt{You are a strict evaluator of startup success predictions.}\\ 
\texttt{Your task is to review an initial prediction and its reasoning, and then provide a final, refined prediction that is logically consistent with the data and decision policy.}\\  \newline 
\texttt{The AI initially predicted: \{initial prediction\} with the following reasoning: \{initial reasoning\}}\\ 
\texttt{Double-check if the reasoning follows the decision policy and all provided data.}\\ 
\texttt{If there are any errors or omissions, correct them.}\\ 
\texttt{Finally, output the final correct outcome as either 'HIGH' or 'LOW', followed by a brief explanation of your corrections.}

\begin{table}[ht]
\centering
\begin{tabular}{lcc}
\toprule
\textbf{Metric} & \textbf{o3 (Single-pass)} & \textbf{o3 (2-step)} \\
\midrule
Precision & 0.225 & 0.237 \\
Recall    & 0.900 & 0.900 \\
F1 Score  & 0.368 & 0.375 \\
MCC       & 0.221 & 0.247 \\
Accuracy  & 0.467 & 0.500 \\
\bottomrule
\end{tabular}
\caption{Performance Comparison: Single-pass o3-mini vs. Two-step Refinement}
\label{tab:o3_vs_2step}
\end{table}

\begin{table}[ht!]
\centering
\begin{tabular}{l|cc}
\multicolumn{1}{c}{} & \multicolumn{2}{c}{\textbf{Predicted}}\\
\cline{2-3}
 & Failure & Success \\
\hline
\textbf{Actual Failure} & 21 & 29 \\
\textbf{Actual Success} & 1 & 9 \\
\end{tabular}
\caption{Confusion Matrix for the o3-mini Model with two-step refinement}
\label{tab:cm2s}
\end{table}

Table~\ref{tab:o3_vs_2step} compares the key performance metrics of the \texttt{o3-mini} model using a single-pass approach versus a two-step refinement process. Notably, the two-step refinement leads to improvements across several metrics. For example, precision increases from 0.225 to 0.237, and the F1 score rises from 0.368 to 0.375. MCC also improves from 0.221 to 0.247, with overall accuracy increasing from 46.7\% to 50.0\%. Although these improvements are small, they are significant in applications where even incremental gains can impact decision-making reliability.

This distribution yields a high recall of 0.900 for successful cases, ensuring that the model captures most true positives. However, the precision is relatively low (approximately 0.237) due to a substantial number of false positives in the failure class. 

Overall, the data suggest that the two-step refinement process enhances the model's performance by providing more reliable and interpretable predictions. Nonetheless, further efforts are warranted to reduce the false positive rate and achieve a more balanced classification performance. 

\subsection{o3 vs simulated reinforcement learning (RL)}

In another set of experiments, we introduced a simulated reinforcement learning (RL) component to serve as a secondary critic of the \texttt{o3-mini} model's reasoning quality. Our framework incorporates a simulated RL scoring mechanism to quantitatively evaluate the quality of the model's chain-of-thought reasoning. This simulated RL step introduces an additional level of quality control by rewarding coherent and precise reasoning while penalising outputs that are vague or logically inconsistent. Specifically, the system prompts the \texttt{o3-mini} model to assign a score between 0 and 1 to each candidate's reasoning, where a score of 1 indicates perfect precision (i.e., no false positives), and a score of 0 denotes completely incorrect reasoning. This base score is then adjusted based on the ground truth of the prediction: when both the actual outcome and the prediction are positive (a true positive), a reward of +0.2 is added; if the actual outcome is negative whilst the prediction is positive (a false positive), a penalty of -0.2 is applied; similarly, a false negative is penalised by -0.1, whereas a true negative receives a modest reward of +0.05. These scores have been tweaked to make the LLM not react too harshly to false positives. 

\begin{figure*}[b!]
    \centering
    \includegraphics[width=0.65\textwidth]{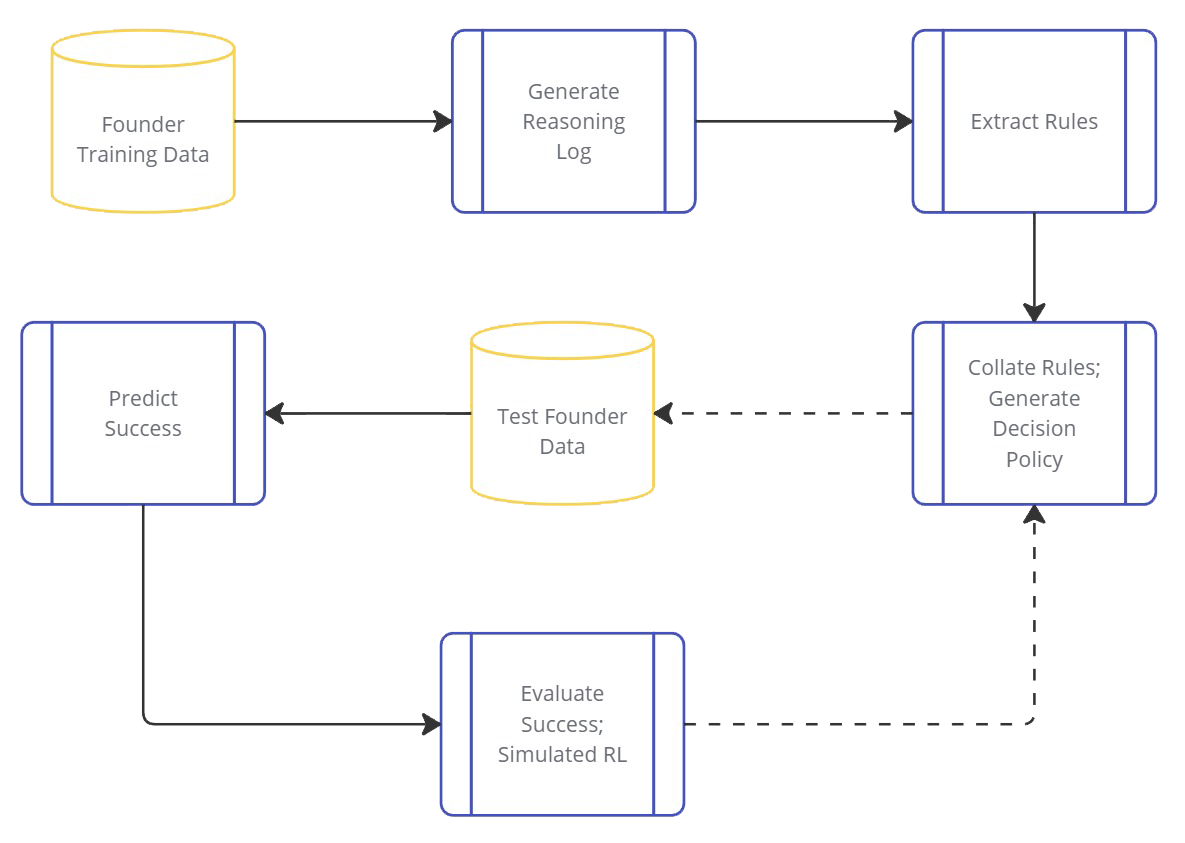}
    \caption{Overview of the Simulated Reinforcement Learning Module used to refine the quality of the model’s reasoning}
    \label{fig:simrl}
\end{figure*}

This score is tied to the rules and is fed back into the refine policy function; this function reads the extracted rules and test predictions. It aggregates the extracted rules into a summary and computes performance data (average RL score, overall accuracy, and sample predictions). It then builds a long text prompt that includes the current decision policy, the rules summary, and test results. This prompt is sent to the model with instructions to refine the decision policy, emphasising precision based on the prior scoring. The refined policy text is returned. This mechanism not only encourages coherent and logically sound explanations but also provides a valuable signal for refining the overall decision policy, ultimately leading to improved prediction accuracy and interpretability. 

An example prompt is as follows:

\noindent\texttt{You are an expert evaluator of startup success decision policies.}\\ 
\texttt{Refine the following decision policy for predicting startup \{category\} outcomes, improving detail, with a strong focus on precision (minimise false positives).}\\ 
\texttt{Integrate all key signals from the extracted rules below without contradicting the data.}\\ 
\texttt{Incorporate insights from our RL-based scoring mechanism: ensure that the refined policy emphasises the importance of high-quality reasoning that earns high RL scores by rewarding true positives and penalising false positives.}\\ 
\texttt{Use this scoring feedback to enhance the robustness and interpretability of the final decision policy.}

In comparison to using \texttt{o3-mini} alone, incorporating simulated RL-based scoring resulted in a measurable improvement in the overall decision quality and interpretability of the extracted rules.

\begin{table}[ht]
\centering
\begin{tabular}{lcc}
\toprule
\textbf{Metric} & \textbf{o3 (No RL)} & \textbf{o3 (Simulated RL)} \\
\midrule
Precision & 0.225 & 0.243 \\
Recall    & 0.900 & 0.900 \\
F1 Score  & 0.368 & 0.383 \\
MCC       & 0.221 & 0.261 \\
Accuracy  & 0.467 & 0.517 \\
\bottomrule
\end{tabular}
\caption{Performance Comparison: o3-mini without vs. with Simulated RL Scoring}
\label{tab:o3_vs_simRL}
\end{table}

\begin{table}[ht!]
\centering
\begin{tabular}{l|cc}
\multicolumn{1}{c}{} & \multicolumn{2}{c}{\textbf{Predicted}}\\
\cline{2-3}
 & Failure & Success \\
\hline
\textbf{Actual Failure} & 17 & 33 \\
\textbf{Actual Success} & 1 & 9 \\
\end{tabular}
\caption{Confusion Matrix for the o3-mini Model Before RL}
\label{tab:cmb4}
\end{table}

\begin{table}[ht!]
\centering
\begin{tabular}{l|cc}
\multicolumn{1}{c}{} & \multicolumn{2}{c}{\textbf{Predicted}}\\
\cline{2-3}
 & Failure & Success \\
\hline
\textbf{Actual Failure} & 22 & 28 \\
\textbf{Actual Success} & 1 & 9 \\
\end{tabular}
\caption{Confusion Matrix for the o3-mini Model After RL}
\label{tab:cmaf}
\end{table}

The data presented in Table~\ref{tab:o3_vs_simRL} and the accompanying confusion matrices indicate that incorporating simulated RL scoring into the \texttt{o3-mini} model leads to measurable performance improvements. With simulated RL, the precision increases from 0.225 to 0.243 and the overall accuracy increases from 46.7\% to 51.7\%, while the recall remains constant at 0.900. These improvements are also reflected in the F1 score and MCC, which increase from 0.368 to 0.383 and from 0.221 to 0.261, respectively. The confusion matrices reveal a notable shift: before applying RL, the model predicted 17 failures and 33 successes for actual failure cases, but after RL, it correctly classifies a higher number of failures (22) and fewer successes (28) for the same group, indicating enhanced discrimination between the two classes. An average RL score of 0.042 from the initial test suggests that there is still room for further optimisation in the RL-based evaluation process. 

This simulated RL mechanism helps to emphasise coherent and logically sound reasoning while penalising vague or overly generic explanations. The introduction of simulated RL resulted in a noticeable improvement in overall decision quality, with the refined scores correlating well with improved prediction accuracy. This approach demonstrates that even without a full RL training loop, incorporating an RL-style evaluation can significantly improve model performance.

The final RL score, obtained by combining the base score with the adjustments, serves as a robust quantitative measure of the quality of reasoning.

\subsection{o3 vs 3 picks}

We further enhanced our prediction mechanism by adopting an ensemble candidate sampling strategy. For each founder, the \texttt{o3-mini} model is prompted three times independently to generate candidate predictions along with their explanations. By applying a simple majority vote over these three candidate predictions, thereby providing a range of perspectives.

The rationale is that individual predictions might vary due to stochastic variations in the model's output and this strategy leverages the natural variability in the model’s output to arrive at a consensus decision. By averaging over three predictions, we reduce the likelihood of and filter out sporadic errors or outlier responses. Our experiments showed that this method increases the stability and reliability of the final prediction compared to relying on a single prediction from \texttt{o3-mini}, leading to improved precision and consistency across the test set.

\begin{figure}[ht!]
    \centering
    \includegraphics[width=0.45\textwidth]{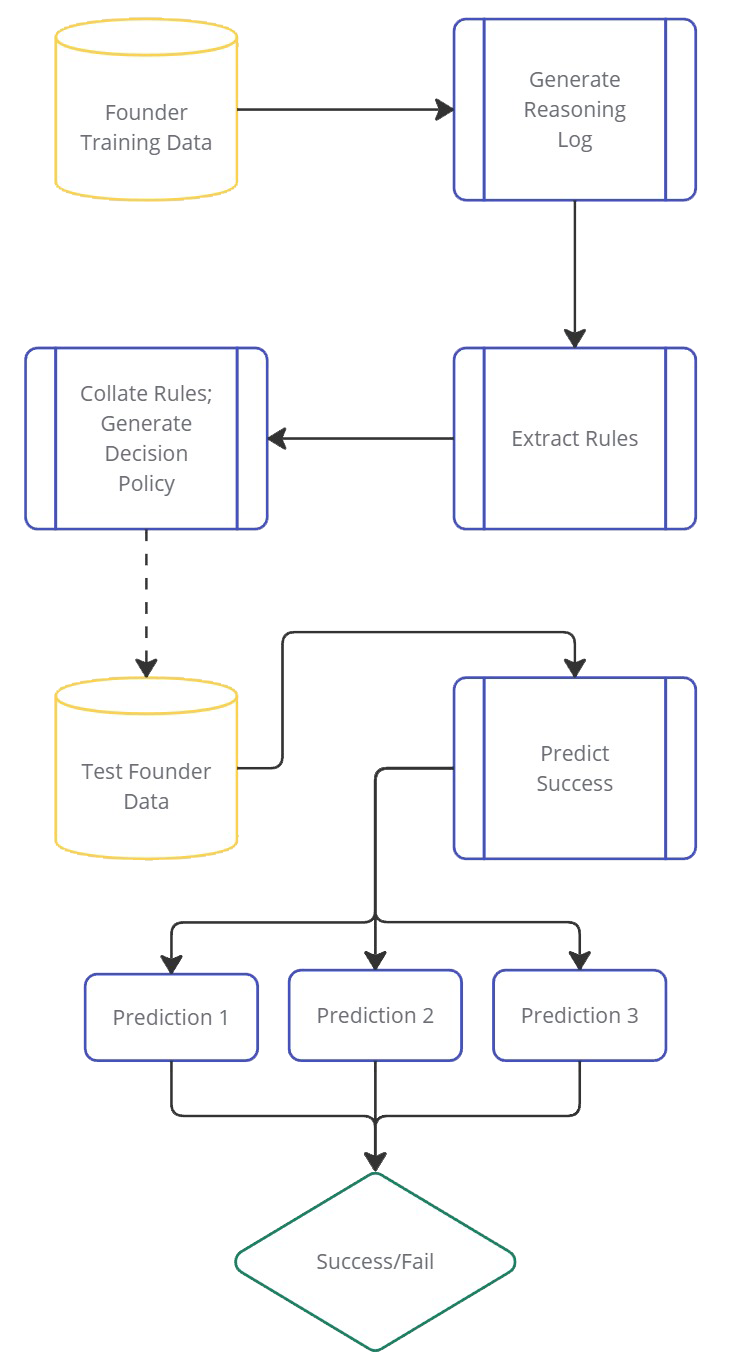}
    \caption{Overview of the Ensemble Candidate Sampling Process}
    \label{fig:3pick}
\end{figure}

\begin{table}[ht!]
\centering
\begin{tabular}{lcc}
\toprule
\textbf{Metric} & \textbf{o3 (Single Prediction)} & \textbf{o3 (3 Picks)} \\
\midrule
Precision & 0.225 & 0.265 \\
Recall    & 0.900 & 0.900 \\
F1 Score  & 0.360 & 0.409 \\
MCC       & 0.221 & 0.301 \\
Accuracy  & 0.467 & 0.567 \\
\bottomrule
\end{tabular}
\caption{Performance Comparison: o3-mini Single Prediction vs. Ensemble of 3 Picks}
\label{tab:o3_vs_3picks}
\end{table}

\begin{table}[ht!]
\centering
\begin{tabular}{l|cc}
\multicolumn{1}{c}{} & \multicolumn{2}{c}{\textbf{Predicted}}\\
\cline{2-3}
 & Failure & Success \\
\hline
\textbf{Actual Failure} & 25 & 25 \\
\textbf{Actual Success} & 1 & 9 \\
\end{tabular}
\caption{Confusion Matrix for the o3-mini Model with Ensemble}
\label{tab:cm3p}
\end{table}

The performance metrics in Table~\ref{tab:o3_vs_3picks} demonstrate that the employment of an ensemble approach with the \texttt{o3-mini} model leads to noticeable improvements over a single prediction strategy. Specifically, the precision increases from 0.225 to 0.265, while the F1 score and MCC improve from 0.360 to 0.409 and from 0.221 to 0.301, respectively. This enhancement is also reflected in the overall accuracy, which rises from 46.7\% to 56.7\%. Notably, recall remains constant at 0.900, indicating that the model’s ability to correctly identify true positives is preserved. The confusion matrix in Table~\ref{tab:cm3p} further highlights these improvements: for actual failures, the model yields an equal number of predictions for failure and success, whereas for actual successes, the ensemble method correctly identifies a larger ratio of actual success. Overall, these results suggest that the aggregation of multiple predictions by a majority vote not only reduces stochastic errors but also enhances the stability and reliability of the model’s performance in classifying startup outcomes. 

\subsection{o3 vs memory}
Persistent conversational memory plays a vital role in our pipeline by allowing the system to retain and leverage context across multiple interactions. By integrating a summary-based memory module with \texttt{o3-mini} as developed by LangChain, our approach captures essential details from prior dialogues, reducing redundant information and token overhead. This not only augments the consistency of the chain-of-thought explanations but also prevents the model from repeating itself, thereby ensuring that each response builds naturally on the conversation history.

\begin{figure*}[b!]
    \centering
    \includegraphics[width=0.75\textwidth]{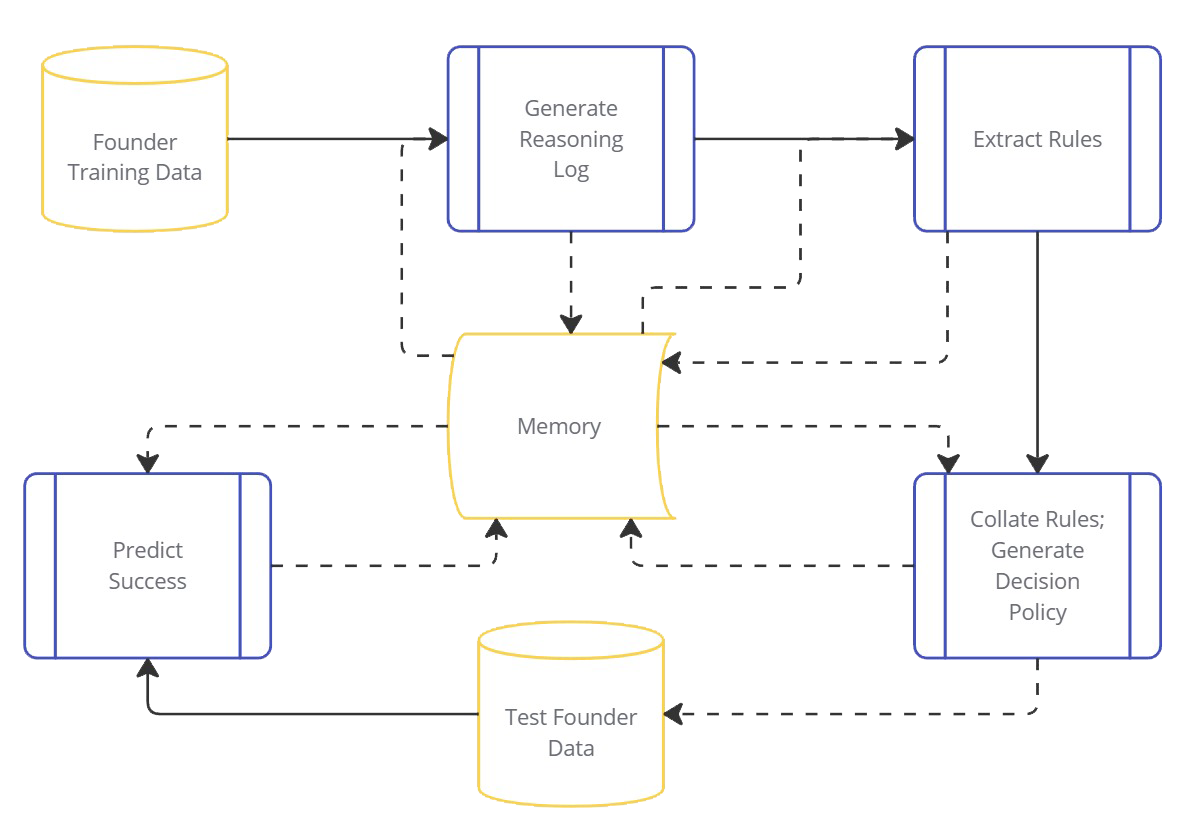}
    \caption{Overview of the Persistent Memory Integration, showing how dynamic summarisation maintains context across multiple interactions.}
    \label{fig:Memory2}
\end{figure*}

\begin{figure*}[t!]
    \centering
    \includegraphics[width=0.8\textwidth]{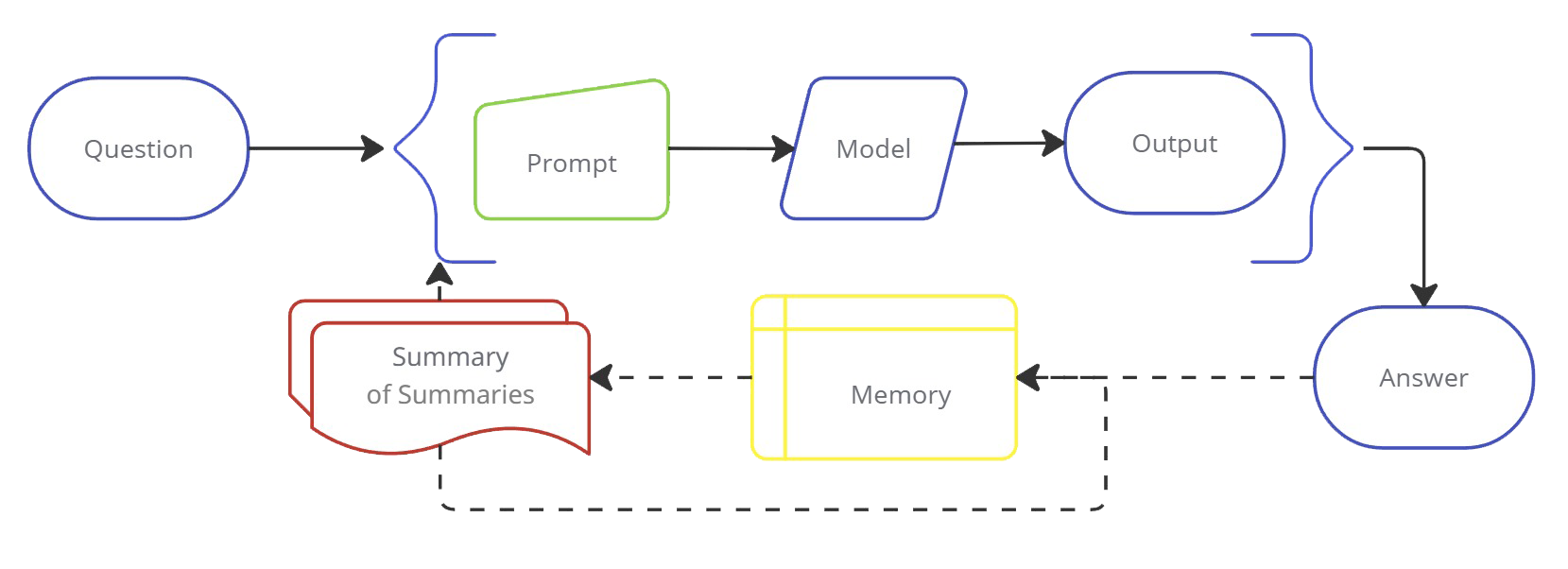}
    \caption{Overview of the Memory Module that supports the entire decision-making process by retaining key insights.}
    \label{fig:Memory1}
\end{figure*}

Our approach leverages dynamic summarisation techniques that continuously compress previous dialogue into succinct summaries, capturing the essential details of past interactions while filtering out unnecessary information. This process ensures that the model focuses on the most relevant context, allowing it to build effectively on earlier reasoning steps. The memory module seamlessly supports every stage of the process, from generating detailed reasoning logs to making informed predictions, thereby optimising the overall performance and precision of our system. By updating and refining the conversation history in this way, the system is better positioned to generate precise and contextually informed predictions. The memory pipeline used is shown in \Cref{fig:Memory1}. This memory is accessed across all steps from reasoning log generation to prediction, as shown in \Cref{fig:Memory2}.

Experimental results demonstrate that models incorporating persistent memory generated more focused and concise reasoning logs compared to those without it and deliver higher-quality predictions. Integrating memory into the process builds upon the accuracy and clarity of our decision policies by ensuring that key insights from earlier interactions inform later decisions. By feeding these refined summaries back into the model, our framework ensures that all critical signals from earlier interactions are incorporated into subsequent decision-making. 

An important consequence of incorporating persistent memory is the production of stronger decision rules. By "stronger," we refer to a set of rules that are more coherent, stable, and reflective of the key signals distilled from prior interactions. The enhanced clarity and focused context help ensure that these rules capture the most important factors influencing decisions, thereby increasing their predictive value. This systematic refinement makes the rule set a more reliable and interpretable foundation for continuous policy evolution.

\begin{table}[ht!]
\centering
\begin{tabular}{lcc}
\toprule
\textbf{Metric} & \textbf{o3 (No Memory)} & \textbf{o3 (With Memory)} \\
\midrule
Precision & 0.225 & 0.321 \\
Recall    & 0.900 & 0.900 \\
F1 Score  & 0.368 & 0.474 \\
MCC       & 0.221 & 0.388 \\
Accuracy  & 0.467 & 0.667 \\
\bottomrule
\end{tabular}
\caption{Performance Comparison: o3-mini without vs. with Persistent Memory}
\label{tab:o3_vs_memory}
\end{table}

\begin{table}[ht!]
\centering
\begin{tabular}{l|cc}
\multicolumn{1}{c}{} & \multicolumn{2}{c}{\textbf{Predicted}}\\
\cline{2-3}
 & Failure & Success \\
\hline
\textbf{Actual Failure} & 31 & 19 \\
\textbf{Actual Success} & 1 & 9 \\
\end{tabular}
\caption{Confusion Matrix for the o3-mini Model with Persistent Memory}
\label{tab:cmmem}
\end{table}

The integration of persistent memory in the \texttt{o3-mini} model yields a notable improvement in performance, as evidenced by Table~\ref{tab:o3_vs_memory}. Specifically, the precision increases from 0.225 to 0.321, which in turn boosts the F1 score from 0.368 to 0.474, and the MCC from 0.221 to 0.388. The confusion matrix presented in Table~\ref{tab:cmmem} further supports these findings: for 50 actual failure cases, 31 are correctly identified as failures while 19 are misclassified as successes, and among 10 actual success cases, 9 are correctly predicted with only 1 misclassified as a failure. Thus overall accuracy rises from 46.7\% to a high score of 66.7\%, while recall remains constant at 0.900. This indicates that the inclusion of persistent memory significantly reduces false positives, thereby enhancing the model's overall prediction reliability.

These results suggest that persistent memory enables the model to retain crucial context across interactions, leading to more precise and interpretable decision-making.

\subsection{o3 vs combined}

Finally, we introduce our end-to-end pipeline for predicting startup success, which brings together several advanced techniques: multistep refinement, ensemble candidate sampling, simulated RL-based scoring, and persistent memory. By combining these methods, our approach leverages the unique strengths of each component to enhance both the accuracy and interpretability of our predictions. The combined approach leverages all the enhancements discussed above. Our results show that the fully integrated pipeline outperforms all individual components. Our method not only builds on traditional LLM outputs but also refines them into a more human-understandable decision policy. It achieves the highest precision and the most interpretable decision policies. 

\begin{figure*}[t]
    \centering
    \includegraphics[width=0.7\textwidth]{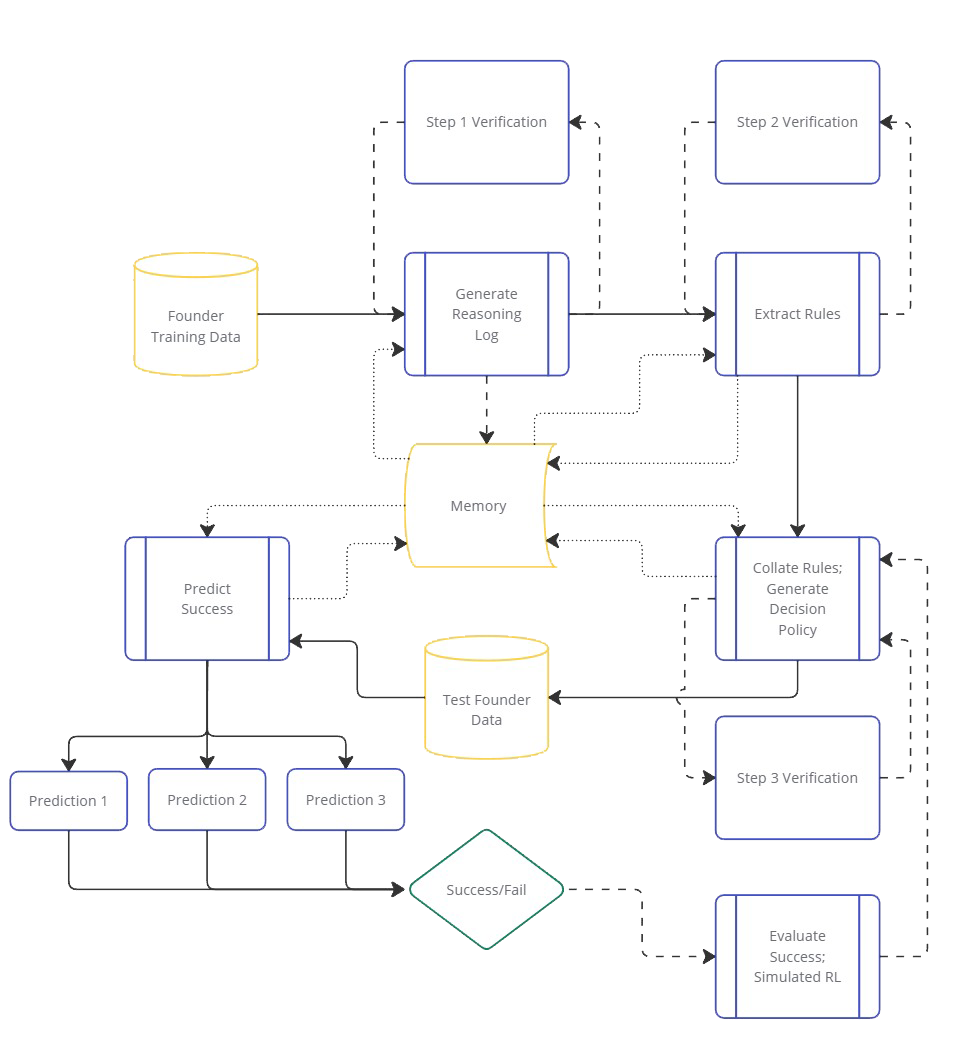}
    \caption{Integrated Pipeline Flowchart, combining multi-step refinement, ensemble sampling, simulated RL scoring, and memory.}
    \label{fig:Combined}
\end{figure*}

Figure~\ref{fig:Combined} provides an overview of the combined pipeline, while Table~\ref{tab:o3_vs_combined} presents a summary of the performance metrics compared to the baseline \texttt{o3-mini} model. Table~\ref{tab:cmo3_combined} shows the corresponding confusion matrix.

The confusion matrix in Table~\ref{tab:cmo3_combined} further reinforces these improvements by showing a marked increase in correctly predicted successful cases. Beyond simply increasing the overall accuracy, our analysis reveals that the combined pipeline significantly reduces the number of false positives in contrast to the baseline \texttt{o3-mini} model; the integrated approach offers enhanced discrimination between successful and failed cases, thereby increasing the trustworthiness of its recommendations since overestimating a startup's potential can lead to substantial financial risk.

\begin{table}[ht]
\centering
\begin{tabular}{lcc}
\toprule
\textbf{Metric} & \textbf{o3-mini Baseline} & \textbf{Combined Pipeline} \\
\midrule
Precision & 0.225 & 0.346 \\
Recall    & 0.900 & 0.900 \\
F1 Score  & 0.368 & 0.500 \\
MCC       & 0.221 & 0.421 \\
Accuracy  & 0.467 & 0.700 \\
\bottomrule
\end{tabular}
\caption{Performance Comparison: o3-mini Baseline vs. Fully Combined Pipeline}
\label{tab:o3_vs_combined}
\end{table}

\begin{table}[ht!]
\centering
\begin{tabular}{l|cc}
\multicolumn{1}{c}{} & \multicolumn{2}{c}{\textbf{Predicted}}\\
\cline{2-3}
 & Failure & Success \\
\hline
\textbf{Actual Success} & 33 & 17 \\
\textbf{Actual Failure} & 1 & 9 \\
\end{tabular}
\caption{Confusion Matrix for the Combined o3-mini Model}
\label{tab:cmo3_combined}
\end{table}

The results clearly demonstrate the benefits of our integrated approach. The fully combined pipeline achieves a precision of 0.346, an F1 score of 0.500, and an MCC of 0.421, with overall accuracy rising from 46.7\% to 70.0\%. These improvements indicate that the combination of multistep refinement, ensemble candidate sampling, RL-based scoring, and persistent memory effectively reduces false positives and increases the reliability of predictions. The synergy among the modules creates a feedback loop as the strengths of one component compensate for the weaknesses of another. The enhanced consistency across multiple evaluation metrics reflects not only improved statistical performance but also a more coherent explanation structure. Finally, the ability of the integrated pipeline to dynamically adjust and refine decision boundaries further cements its superiority over isolated approaches.

Moreover, the confusion matrix confirms that our holistic framework is more adept at distinguishing between successful and failed cases. In particular, there is a significant increase in the correct identification of successful startups. These findings suggest that our approach not only produces more accurate predictions, but also produces decision policies that are easier for human experts to interpret and trust. In general, the integration of these advanced techniques creates an interpretable system that holds great promise for improving startup success prediction.

\section{Discussion and Future Work}

Our proposed framework introduces a novel and interpretable approach to evaluating startup founders by combining LLM-based reasoning with structured, rule-based decision-making. Unlike traditional machine learning models, which often function as opaque predictors, our method ensures that each prediction can be traced back to explicit, human-understandable reasoning. The system generates clear, natural-language explanations alongside each prediction, allowing decision-makers to trace outcomes back to specific, human-understandable rules. Every founder classified as “Successful” is accompanied by a natural-language explanation or a rule (e.g., “Founder has a prior exit and recent funding success”) that justifies the decision. Its modular design, which includes multistep refinement, ensemble candidate sampling, simulated RL-based scoring, and persistent memory, enables stable performance while maintaining transparency. This comprehensive approach not only automates complex evaluations but also facilitates expert intervention, allowing domain experts to review and refine the distilled policies to better capture industry insights.

While our framework shows promising improvements in prediction accuracy and interpretability, several opportunities remain for further enhancement. Future work should explore deeper reasoning verification, incorporate more dynamic human-in-the-loop refinements, and test the approach across a broader range of domains. For instance, increasing the number of candidate predictions (samples) might boost performance, yet it raises the question of whether comparing and retaining only the best-performing samples or a top-N selection could refine the outcomes any further. Additionally, converting the extracted rules back into natural language (for example, consolidating them into an “investment thesis”) could improve usability for end users by presenting a coherent and comprehensive explanation of the decision policy. These advances, including the development of secondary verification mechanisms and adaptive policy documentation, will be crucial for ensuring that the model remains stable in the face of out-of-distribution data and evolving market conditions.

Key advantages of our framework include its modular and iterative design, which enables the following:
\begin{itemize}
    \item \textbf{Diverse reasoning perspectives:} By sampling multiple chain-of-thought outputs and using memory to accumulate insights, the single LLM effectively simulates a panel of judges with varied viewpoints. This ``crowd of one'' approach takes inspiration from the wisdom-of-crowds effect in LLM ensembles, yielding more accurate decisions than a single pass.
    \item \textbf{Self-optimisation through feedback:} We demonstrated a process by which the LLM can refine its own decision policy using performance feedback, without additional training data. In this process, the model essentially "learns" from its immediate mistakes: it observes which predictions were less accurate, refines its internal logic accordingly, and thereby improves its chain-of-thought reasoning. This dynamic adjustment is a form of 'in-context' learning and policy self-improvement that mimics test-time optimisation, where the model continuously fine-tunes its behaviour during inference to achieve higher precision.
    
    \item \textbf{Flexible prompts and domain adaptation:} Our framework is inherently adaptable, allowing targeted prompts to incorporate domain-specific data and expert insights. This flexibility ensures that the system can evolve with changing market conditions and tailor its decision policies to meet the unique challenges of different industries. 
    
\end{itemize}

Despite these advantages, several areas warrant further investigation:
\begin{itemize}
    \item \textbf{Incorporating human-edited policies:} While our LLM autonomously refined the rule set, a human-in-the-loop approach could refine them further. Future studies should examine how direct domain expert adjustments to the rules impact performance and whether such curated policies generalise better or avoid certain biases. 
    
    This ensures that the explanations not only make sense on paper but also translate into better, more informed investment outcomes in real-world settings. Engaging venture capitalists and investment analysts in controlled experiments and real-life pilot programs can provide valuable feedback on the clarity, usefulness, and reliability of the AI-generated insights, allowing researchers to fine-tune the system's reasoning and presentation methods based on practical needs. Moreover, such evaluations can reveal whether the model's transparent decision-making process helps experts identify previously overlooked risks or opportunities.
    \item \textbf{Cost and token use:} A drawback of using the additional modules is that we are calling the model more frequently, thereby increasing token use and cost substantially---2× for the 2-step process, 3× for the ensemble, and 8× for the combined model. Since the improvement from combining methods over using memory alone is minimal, future work focused on the memory-only approach could be highly beneficial. Fewer API calls will be preferable. 

    \item \textbf{Enhancing reasoning depth and verification:} Our framework already uses chain-of-thought prompting to encourage multi-step reasoning, but there remains significant room to deepen the model's logical engagement with difficult cases. One promising technique is \emph{budget forcing}, which combats premature termination of reasoning by appending a self-interrogation prompt—such as “Wait? Did I make a mistake?”—at the end of each intermediate response. This discourages the model from treating its answer as final and instead prompts it to continue reflecting. The follow-up user message instructs the model to “Continue your thinking where you left off, correcting any mistakes if there is any. Think for up to 8096 tokens,” guiding the model through iterative refinement cycles. After several such steps, the process is concluded with an instruction to provide a concise, final answer (e.g., “Respond in under 2048 tokens!!!”). This structured loop encourages the model to revisit and improve its prior reasoning, often leading to clearer logic and reduced error rates. By simulating deeper critical thinking, budget forcing enforces both accuracy and the coherence of explanations—particularly in edge cases where reasoning quality typically suffers. That said, overuse can introduce unnecessary verbosity or new errors, so tuning the number of iterations—potentially adjusting dynamically based on difficulty—remains an important area for future exploration.

    \item \textbf{Evaluating stability across domains:} We focused on startup founder success, but the method can be applied to other domains that require interpretable decision-making (hiring decisions, grant selections, etc.). It would be valuable to test if the same LLM reasoning approach and iterative refinement yield similar gains in those contexts, or if domain-specific tuning is required. Additionally, stability tests (e.g., how the model handles out-of-distribution founder profiles or adversarially crafted profiles) are needed to ensure reliability.
    \item \textbf{Adaptive policy documentation:} To maximise real-world usability, the final set of decision rules could be compiled into a formal policy document for VC firms. We aim to automate the production of such a document, where each rule is accompanied by a rationale and statistics (e.g., “Rule~1 correctly identified 80\% of successful founders with no false positives in testing”). Such a document would serve as a transparent and editable investment framework. Periodically, as new data is received, the system could update this policy, essentially becoming a living documentation that evolves with the market.
    \item \textbf{Rules as code:} A possible extension is to output rules as executable code—akin to CodeAct or similar methods. This would allow immediate validation via a compiler or interpreter, letting one test the rules against real founder data. Such a workflow not only ensures logical and syntactic soundness but also streamlines integration into production pipelines where validated, executable logic is essential.
    \item \textbf{Improved hallucination detection:} By integrating hallucination detection libraries into our workflow, it would be possible to monitor and evaluate LLM outputs for potential hallucinations. This system would provide binary scores that indicate the likelihood of a response being hallucinated, thereby adding an extra layer of reliability and trustworthiness to the predictions.

    \item \textbf{Fine-tuning the LLM:} We conducted initial fine-tuning calls on the \texttt{o3-mini} model to better adapt its reasoning to the domain of startup evaluation. However, there remains significant scope for further improvement through more targeted fine-tuning using larger, domain-specific datasets and iterative calibration. Enhancing the model's ability to generate coherent and precise explanations could further improve both prediction accuracy and the quality of the extracted decision rules.
\end{itemize}

By refining these components, we can further enhance LLM-powered and interpretable investment models. Our work bridges the gap between the predictive prowess of LLMs and the practical needs of decision-makers for understandable and controllable models, and we see promising avenues for extending this paradigm to support human decision-makers in other high-stakes, uncertain environments.

\section{Conclusion}
We propose a hybrid LLM-driven reasoning system that enables explainable investment decision-making. Our approach demonstrates strong predictive power using interpretable heuristics extracted from an LLM's chain-of-thought, achieving substantially higher precision than a purely random or baseline model. Importantly, each prediction is accompanied by a clear text-based explanation, and the decision policy is expressed as an editable list of rules. This framework shows promise for integration into real-world VC workflows, where it can augment human investors by providing data-driven insights that are immediately understandable. In summary, our results highlight that large language models, when guided to reason and explicate their reasoning, can be powerful allies in complex decision domains, combining the flexibility of AI with the transparency of rule-based systems.

Our experimental evaluations on curated startup datasets demonstrated that each component of our framework contributes to a more stable and reliable decision-making process. The integration of persistent memory and ensemble methods significantly reduced false positives. The two-step refinement and RL-based scoring further improved the quality of the reasoning logs. The combined pipeline achieved notable gains in precision, F1 score and overall accuracy compared to baseline models, underscoring the potential of our framework in high-stakes investment environments.

Despite these promising results, our framework also faces several challenges. The increased token use and computational cost associated with multistep and ensemble approaches highlight the need for further optimisation, particularly when scaling the model to larger datasets or real-time applications. Moreover, while our system bolsters transparency, ensuring that the generated explanations are accurate and free from hallucinations remains an ongoing research focus. 

Future work should explore more efficient memory management strategies, advanced fine-tuning on domain-specific datasets, and targetted hallucination detection techniques to further refine decision policies. The expansion of the framework to other domains such as healthcare, finance, or hiring could unlock new opportunities for explainable AI in complex decision-making scenarios. Ultimately, our research contributes to the growing body of work that seeks to integrate human expertise with AI-driven insights, laying the groundwork for more transparent and effective investment decision frameworks.

\bibliographystyle{plainnat}

\bibliography{main}

\section*{Impact Statement}
Our framework harnesses the advanced reasoning capabilities of large language models to deliver interpretable and actionable investment predictions, addressing a critical need in the venture capital domain. By transforming LLM-generated reasoning logs into explicit and verifiable decision rules, the system empowers investors to understand and refine the decision-making process. This transparency not only builds trust in AI-driven recommendations but also facilitates expert intervention, ultimately reducing investment risk and uncertainty by clarifying investment decisions, emerging investment rules, and the investment thesis.

Beyond venture capital, this approach has the potential to revolutionise decision-making in other high-stakes fields like healthcare and finance, where explainable AI is essential for informed, accountable and reliable outcomes.

\quad %
\newpage

\appendix
\appendixpage

\section{4o-mini Reasoning Logs, Extracted Rules, and Final Decision Policy}

This appendix presents Anonymised reasoning logs for both successful and failed founder cases evaluated using GPT 4o, along with the extracted decision rules and final decision policies.

\subsection{Anonymised Reasoning Logs}

\subsubsection{Success Cases}

\textbf{Founder A - Success}\\ 
\textbf{Key Reasons for Founder Success: Founder A}
\begin{enumerate}
    \item \texttt{Extensive Industry and Operational Experience: Founder A has over 20 years of experience in technology, product development, management, and strategy. Previous roles, including CEO of a major company (acquired for \$360M) and senior positions at leading technology firms, provided deep insights into operational best practices and technical innovation.}
    \item \texttt{Proven Track Record: The successful acquisition of the former company validates Founder A's capability to build and scale a technology-driven enterprise, enhancing their credibility with investors and strategic partners.}
    \item \texttt{Strong Dual Expertise in Technology and Business: Holding dual degrees in Software Engineering and Business, Founder A effectively bridges technical innovation and sound business strategy, a critical combination for startups leveraging technology to disrupt traditional markets.}
    \item \texttt{Strategic Vision: Their experience in corporate development and mergers and acquisitions equips Founder A with the strategic foresight necessary for scaling a startup and capturing market opportunities.}
    \item \texttt{Focus on Technology and Innovation: The startup leverages cutting-edge technology to streamline processes and drive innovation, ensuring a competitive edge in the market.}
    \item \texttt{Agility and Adaptability: The operational framework emphasizes quick decision-making and responsiveness to market dynamics.}
    \item \texttt{Robust Professional Network: An extensive network of industry connections provides Founder A with mentorship, strategic partnerships, and access to crucial funding.}
\end{enumerate}
\textbf{Conclusion:}\\
\texttt{Founder A's success is primarily driven by their extensive industry experience, proven track record, balanced technical and business expertise, and strong professional networking.}

\vspace{1em}
\textbf{Founder B - Success}\\ 
\textbf{Key Reasons for Founder Success: Founder B}
\begin{enumerate}
    \item \texttt{Deep Financial Services Expertise: Founder B’s long-standing background in financial services gives them a profound understanding of market dynamics, regulatory challenges, and customer needs.}
    \item \texttt{Exceptional Communication and Investor Relations Skills: Their experience in executive roles focused on communications and investor relations has enabled effective articulation of the startup’s value proposition.}
    \item \texttt{Strategic Networking: With a robust professional network, Founder B has successfully established strategic partnerships and secured investor confidence.}
    \item \texttt{Innovative Product Offering: The startup provides a user-friendly, comprehensive solution in digital asset management, addressing a growing market need.}
    \item \texttt{Clear Leadership and Vision: Founder B’s ability to align team and stakeholder objectives through clear strategic vision is a key factor in early market success.}
    \item \texttt{Market Timing and Regional Alignment: Timely market entry, combined with geographic alignment in a dynamic financial hub, has further propelled the startup’s growth.}
    \item \texttt{Focus on Trust and Security: Emphasizing security in transactions has reinforced customer and institutional confidence.}
\end{enumerate}
\textbf{Conclusion:}\\
\texttt{Founder B’s success is attributed to extensive financial expertise, clear leadership, strategic networking, and a well-timed market entry.}

\subsubsection{Failure Cases}

\textbf{Founder C - Failure}\\ 
\textbf{Key Reasons for Failure: Founder C}
\begin{enumerate}
    \item \texttt{Product–Market and Technology Mismatch: The startup's flagship product, designed to enhance a specific procedure, failed to clearly outperform established solutions, leading to poor market differentiation.}
    \item \texttt{Regulatory and Operational Challenges: Difficulties in achieving clinical validation, obtaining regulatory approvals, and streamlining operations impeded successful market entry.}
    \item \texttt{Funding Constraints: Insufficient funding restricted the startup's ability to scale product development and optimize go-to-market strategies.}
    \item \texttt{Leadership and Team Dynamics Issues: Ineffective team cohesion and leadership contributed to suboptimal execution of strategic initiatives.}
    \item \texttt{Inadequate Marketing Strategy: Poor market positioning and communication limited product adoption among target customers.}
\end{enumerate}
\textbf{Conclusion:}\\
\texttt{Founder C's failure is primarily due to a lack of clear product-market fit, regulatory and operational obstacles, funding limitations, and leadership challenges.}

\vspace{1em}
\subsubsection{Failure Cases}

\textbf{Founder D - Failure}\\ 
\textbf{Key Reasons for Failure: Founder D}
\begin{enumerate}
    \item \texttt{Lack of Professional Experience: The founder lacks documented professional background or prior entrepreneurial ventures, hindering credibility.}
    \item \texttt{Insufficient Domain Expertise: A limited understanding of the target market's needs resulted in a product that failed to align with consumer expectations.}
    \item \texttt{Limited Network and Resource Access: The absence of a robust professional network significantly impeded access to funding and strategic mentorship.}
    \item \texttt{Poor Product Validation and Adaptation: Failure to validate the product concept through pilot testing resulted in misaligned strategy and execution.}
\end{enumerate}
\textbf{Conclusion:}\\
\texttt{These failure cases emphasize the necessity of deep domain expertise, practical experience, strong networks, and effective product validation for achieving startup success.}

\subsection{Extracted Decision Rules}

The following rules were automatically extracted from the GPT 4o reasoning logs.

\subsubsection{Success Rules}
\begin{itemize}
    \item \textbf{Rule for Founder A:}\\
    \texttt{IF founder has extensive industry and operational experience (including leadership roles and a successful startup exit), dual expertise in technology and business, proven strategic leadership with a clear vision for market disruption, AND a robust professional network boosting credibility THEN likelihood\_of\_success = HIGH.}
    \item \textbf{Rule for Founder B:}\\
    \texttt{IF founder has an extensive background in financial services, demonstrated strong communication and investor relations skills, possesses a strategic professional network, and operates in a region with strong market alignment THEN likelihood\_of\_success = HIGH.}
\end{itemize}

\subsubsection{Failure Rules}
\begin{itemize}
    \item \textbf{Rule for Founder C:}\\
    \texttt{IF founder has a strong corporate industry background BUT does not demonstrate the agile startup execution skills needed to quickly validate product benefits, navigate regulatory hurdles, and drive market adoption THEN likelihood\_of\_success = LOW.}
    \item \textbf{Rule for Additional Reflection on Failure:}\\
    \texttt{IF founder has no documented professional experience AND lacks relevant domain-specific expertise AND possesses a limited professional network AND fails to validate product concepts through feedback THEN likelihood\_of\_success = LOW.}
\end{itemize}

\subsection{Final Decision Policy}

\subsubsection{Success Decision Policy}
Based on the extracted rules from successful founder profiles, here is a concise decision policy summarizing the key conditions that predict startup success:
\begin{enumerate}
    \item \texttt{IF founder has a strong educational background in a relevant field AND extensive industry experience THEN likelihood\_of\_success = HIGH.}
    \item \texttt{IF founder offers an innovative product that addresses a significant market need AND demonstrates a clear vision for growth THEN likelihood\_of\_success = HIGH.}
    \item \texttt{IF founder demonstrates effective leadership skills AND has experience in managing diverse teams THEN likelihood\_of\_success = HIGH.}
    \item \texttt{IF founder possesses a robust professional network AND engages in strategic partnerships THEN likelihood\_of\_success = HIGH.}
    \item \texttt{IF founder demonstrates adaptability to market changes AND resilience in overcoming challenges THEN likelihood\_of\_success = HIGH.}
    \item \texttt{IF founder has previous entrepreneurial experience with a proven track record of success THEN likelihood\_of\_success = HIGH.}
    \item \texttt{IF founder has a strategic vision aligned with market demands AND focuses on innovation THEN likelihood\_of\_success = HIGH.}
    \item \texttt{IF founder shows a commitment to quality products AND social responsibility THEN likelihood\_of\_success = HIGH.}
    \item \texttt{IF founder targets a growing market with a scalable business model AND demonstrates effective marketing strategies THEN likelihood\_of\_success = HIGH.}
    \item \texttt{IF founder has strong technical expertise in relevant technologies AND offers innovative solutions THEN likelihood\_of\_success = HIGH.}
\end{enumerate}

\subsubsection{Failure Decision Policy}
\textbf{Decision Policy for Predicting Startup Failure:}
\begin{enumerate}
    \item \texttt{IF founder has no relevant industry experience AND lacks documented professional experience AND has a limited network AND insufficient market research AND faces operational challenges AND lacks financial backing THEN likelihood\_of\_success = LOW.}
    \item \texttt{IF founder lacks relevant experience in the target market AND focuses on a niche product without broad appeal AND lacks strategic industry connections AND is involved in multiple ventures diluting focus AND faces product development challenges AND has an ineffective marketing strategy THEN likelihood\_of\_success = LOW.}
    \item \texttt{IF founder has extensive experience in a field but lacks business acumen AND faces significant competition AND struggles with operational scalability AND encounters financial management issues THEN likelihood\_of\_success = LOW.}
    \item \texttt{IF founder has no prior entrepreneurial ventures AND operates in a highly competitive market AND lacks a strong marketing strategy AND struggles with operational challenges AND has inadequate funding THEN likelihood\_of\_success = LOW.}
    \item \texttt{IF founder has a strong educational background but lacks experience in entrepreneurship AND faces significant market readiness challenges AND struggles to secure adequate funding AND contends with high competition AND regulatory hurdles THEN likelihood\_of\_success = LOW.}
    \item \texttt{IF founder has extensive experience in a specific field but lacks a clear value proposition AND faces market saturation AND struggles with execution AND has limited marketing strategies AND operates in a sensitive economic environment THEN likelihood\_of\_success = LOW.}
    \item \texttt{IF founder has a diverse skill set but lacks focus on core competencies AND startup faces intense competition without clear differentiation AND struggles with operational management AND financial discipline THEN likelihood\_of\_success = LOW.}
    \item \texttt{IF founder has no relevant experience in the industry AND lacks a clear business model AND faces financial constraints AND struggles with operational execution AND has limited visibility and credibility THEN likelihood\_of\_success = LOW.}
    \item \texttt{IF founder has strong technical expertise but lacks a well-rounded team in sales and marketing AND targets a misaligned market segment THEN likelihood\_of\_success = LOW.}
    \item \texttt{IF founder has limited startup experience AND previous ventures did not achieve notable success AND operates in a highly competitive market with established players THEN likelihood\_of\_success = LOW.}
\end{enumerate}
\texttt{Return: LOW}

\section{o3-mini Reasoning Logs, Extracted Rules, and Final Decision Policy}

This section presents the o3-mini model's reasoning logs for evaluating startup founders, along with the extracted decision rules and the final decision policy. 
\subsection{o3-mini Reasoning Logs}

\subsubsection{Founder A - Success}
\textbf{Key reasons for Founder A's success include:}
\begin{enumerate}
    \item \texttt{Extensive Industry and Operational Experience: Founder A has over 20 years of experience in technology, product development, management, and strategy. Their career includes leadership roles at major companies (e.g., Company X, Company Y) that provided solid operational and technical expertise. Their tenure as CEO at a startup—with a successful acquisition for \$360M—demonstrates the ability to build, scale, and exit a technology-driven company.}
    \item \texttt{Strong Dual Expertise in Technology and Business: Holding degrees in software engineering and business, Founder A effectively bridges the gap between technical innovation and sound business strategy, which is crucial for startups that leverage technology to streamline complex processes. This background ensures the ability to design scalable, user-centric platforms while also managing critical business operations such as raising capital and forming strategic partnerships.}
    \item \texttt{Proven Leadership and Strategic Vision: Their experience in R\&D leadership and corporate development—including managing mergers and acquisitions—provides strong strategic planning capabilities. Their aptitude for identifying market needs and disrupting traditional industries is evident through innovative service models.}
    \item \texttt{Credibility and Robust Network: Having led successful ventures, Founder A has built a reputable profile and an extensive professional network that fosters investor trust and attracts top talent.}
\end{enumerate}
\textbf{Conclusion:}\\
\texttt{Founder A's deep industry experience, dual technical and business acumen, proven leadership, and robust network are key drivers of their success.}

\vspace{1em}
\subsubsection{Founder B - Success}
\textbf{Key reasons for Founder B's success include:}
\begin{enumerate}
    \item \texttt{Deep Financial Services Expertise: Founder B's extensive career in financial services—spanning roles at major firms—provided a profound understanding of market dynamics, investor relations, and corporate communications. This expertise helped establish credibility for their digital asset platform.}
    \item \texttt{Robust Communication and Investor Relations Skills: Their leadership experience as Chief Communications \& Marketing Officer and Head of Investor Relations enabled effective articulation of the platform's value proposition to various stakeholders, including institutional partners.}
    \item \texttt{Strategic Network and Industry Connections: With over 700 professional connections and board involvement, Founder B has leveraged a strong network to secure funding opportunities and strategic partnerships.}
    \item \texttt{Geographic and Sector Alignment: Their regional familiarity and alignment with local financial hubs have contributed to positioning the startup effectively within a growing market focused on digital asset management.}
\end{enumerate}
\textbf{Conclusion:}\\
\texttt{Founder B's combination of deep industry expertise, exceptional communication skills, strategic networking, and market alignment are central to their startup's success.}

\vspace{1em}
\subsubsection{Founder C - Failure}
\textbf{Key reasons for the startup's failure include:}
\begin{enumerate}
    \item \texttt{Product–Market and Technology Mismatch: Despite a strong vision behind the product, it did not deliver clear clinical benefits over conventional imaging methods, failing to integrate into existing workflows.}
    \item \texttt{Regulatory and Validation Hurdles: The product required rigorous clinical validation and regulatory approvals; delays or challenges in demonstrating safety and efficacy hindered market entry.}
    \item \texttt{Execution Challenges in a High-Stakes Environment: Transitioning from a corporate role to a startup, the execution plan may have lacked the agility needed to integrate cutting-edge technology into surgical environments.}
    \item \texttt{Market Adoption Barriers: Without clearly demonstrated improvements in outcomes, the product struggled to secure widespread adoption among experienced users, negatively impacting revenue and investor confidence.}
\end{enumerate}
\textbf{Conclusion:}\\
\texttt{Although the founder possessed deep industry expertise, challenges in proving clinical value, navigating regulatory hurdles, and achieving rapid market adoption contributed to the failure.}

\vspace{1em}
\subsubsection{Founder D - Failure}
\textbf{Key reasons for failure stem from the founder's lack of a robust background:}
\begin{enumerate}
    \item \texttt{Minimal Professional Experience: The founder has no documented professional experience, degrees, or previous entrepreneurial ventures, limiting credibility with stakeholders.}
    \item \texttt{Lack of Domain Expertise: Without demonstrable knowledge in both medical and entrepreneurial domains, the founder struggled to navigate regulatory requirements and secure validation.}
    \item \texttt{Limited Networking and Credibility: The absence of event appearances, press references, or organizational leadership hindered access to mentorship and strategic partnerships.}
    \item \texttt{Inadequate Validation and Feedback: Failure to pilot test and validate the product concept contributed to a misalignment between the offering and market needs.}
\end{enumerate}
\textbf{Conclusion:}\\
\texttt{The founder’s insufficient professional and domain-specific credentials likely led to difficulties in funding, product validation, and execution, ultimately resulting in failure.}

\subsection{Extracted Decision Rules}

Based on the reasoning logs above, the following decision rules were extracted:

\subsubsection{Success Rules}
\begin{itemize}
    \item \textbf{Rule for Founder A:}\\
    \texttt{IF founder has extensive industry and operational experience (including leadership roles and a successful startup exit), dual expertise in technology and business, proven strategic leadership with a clear vision for market disruption, AND a robust professional network boosting credibility THEN likelihood\_of\_success = HIGH.}
    \item \textbf{Rule for Founder B:}\\
    \texttt{IF founder has an extensive background in financial services, demonstrated strong communication and investor relations skills, possesses a strategic professional network, and operates in a region with strong market alignment THEN likelihood\_of\_success = HIGH.}
\end{itemize}

\subsubsection{Failure Rules}
\begin{itemize}
    \item \textbf{Rule for Founder C:}\\
    \texttt{IF founder has a strong corporate industry background BUT does not demonstrate the agile startup execution skills needed to quickly validate clinical benefits, navigate regulatory hurdles, and drive market adoption THEN likelihood\_of\_success = LOW.}
    \item \textbf{Rule for Founder D:}\\
    \texttt{IF founder lacks professional credentials AND lacks relevant medical and entrepreneurial expertise AND has limited industry networking THEN likelihood\_of\_success = LOW.}
\end{itemize}

\subsection{Final Decision Policy}

\subsubsection{Success Decision Policy}
\texttt{IF a founder demonstrates robust, cross-disciplinary academic excellence (e.g., advanced degrees in STEM, business, law, or medicine) AND shows proven domain expertise through hands-on experience in early R\&D, innovation, and product development that leads to measurable impact AND exhibits visionary leadership by effectively managing teams, scaling operations, and making strategic decisions that align technological innovation with clear market needs AND identifies and addresses critical market gaps with innovative, scalable solutions AND leverages a deep, diverse, and active professional network to secure strategic partnerships, mentorship, and capital AND integrates complementary skills (including financial, regulatory, and operational acumen) to navigate complex industry challenges across both emerging and traditional sectors THEN likelihood\_of\_success = HIGH.}

\subsubsection{Failure Decision Policy}
\textbf{Decision Policy for Predicting Startup Failure:}
\texttt{IF the founder’s background and expertise are misaligned with the startup’s specific domain requirements (e.g., coming from non-agile, overly structured, or non-technical roles) OR the startup pursues an overly ambitious or unfocused product vision that dilutes its core value proposition AND faces significant execution challenges – including inability to integrate or scale complex technologies in regulated or legacy environments, resource constraints, weak operational planning, and a lack of complementary, domain-specific leadership and strategic partnerships – AND the overall business strategy is misaligned with the founder’s core skills relative to the startup’s need for rapid, technical innovation and agile market adaptation THEN likelihood\_of\_success = LOW.}

\bigskip

\section{o3 Combined Pipeline. Reasoning Logs and Decision Policies}

\subsubsection{Before Refinement}

\paragraph{Founder A - Success}  
\textbf{Raw Reasoning:} \\
\texttt{
Let's break down the key reasons why Founder A succeeded step by step:
\begin{enumerate}
    \item \textbf{Dual Expertise in Technology and Business:} Founder A’s educational background in both Software Engineering and Business provided a unique blend of technical know‐how and commercial acumen. This dual expertise enabled them to understand intricate product development details while keeping a keen eye on overall business strategy and market trends.
    \item \textbf{Extensive Experience Across the Tech Landscape:} With over 20 years of experience in technology, product development, and management, Founder A accumulated deep practical knowledge from roles at leading companies. This experience sharpened their R\&D skills and the ability to anticipate market needs.
    \item \textbf{Successful Entrepreneurial Track Record:} By co‐founding and leading a company that achieved a significant exit (e.g., a \$360M acquisition), Founder A demonstrated the ability to build and scale a startup, thereby enhancing credibility.
    \item \textbf{Proven Leadership and Strategic Vision:} Experience in executive and corporate development roles enabled Founder A to adopt a strategic outlook, driving large‐scale innovations and timely operational pivots.
    \item \textbf{Strong Operational and Management Skills:} The founder’s hands‐on involvement in managing teams and projects ensured agile execution and effective decision-making.
    \item \textbf{Discipline and High‐Pressure Decision Making:} Unique experiences, such as military service, provided the founder with resilience and the capacity to make calm, informed decisions under pressure.
    \item \textbf{Early Recognition of Technological Opportunities:} Launching a startup that leverages technology to transform traditional processes underlines their ability to seize emerging market opportunities.
\end{enumerate}
}
\textbf{Conclusion:} \\
\texttt{Founder A’s success is driven by a comprehensive integration of technical expertise, extensive experience, proven entrepreneurial success, and visionary leadership.}

\vspace{1em}
\paragraph{Founder B - Success}  
\textbf{Raw Reasoning:} \\
\texttt{
Below is the step-by-step explanation for why Founder B succeeded:
\begin{enumerate}
    \item \textbf{Strong Educational Foundation:} Founder B’s MBA and bachelor's degree in business provided the rigorous training needed in strategy, financial analysis, and management.
    \item \textbf{Deep Experience in Financial Services:} Their career, spanning roles from Equity Research Associate to executive leadership positions, offered deep insights into market dynamics and investor relations.
    \item \textbf{Leadership in Investor Relations and Communications:} Experience in high-profile communication roles built the credibility necessary to secure strategic partnerships.
    \item \textbf{Diverse Industry Exposure and Network Building:} Exposure across multiple sectors resulted in a robust professional network and enriched strategic insights.
    \item \textbf{Strategic Market Alignment:} Effective positioning in the digital asset market reassured investors and customers about the startup’s long-term viability.
    \item \textbf{Resilience and Adaptability:} Proven ability to operate in dynamic, high-pressure environments underscored their capacity to navigate uncertainties.
\end{enumerate}
}
\textbf{Conclusion:} \\
\texttt{Founder B’s success is attributed to a solid educational and professional foundation, effective communication, and a broad, strategically aligned network.}

\vspace{1em}
\paragraph{Founder C - Failure}  
\textbf{Raw Reasoning:} \\
\texttt{
Below is the explanation outlining why Founder C’s venture failed:
\begin{enumerate}
    \item \textbf{Product–Market and Technology Mismatch:} Although the product was innovative, it failed to clearly demonstrate clinical benefits beyond established solutions, resulting in poor market differentiation.
    \item \textbf{Regulatory and Operational Challenges:} Lengthy regulatory processes and difficulties in scaling operations significantly hindered market entry and user adoption.
    \item \textbf{Funding and Resource Constraints:} Inadequate funding limited the startup's capacity to scale product development and commercial execution.
    \item \textbf{Leadership and Execution Issues:} Challenges in team cohesion and strategic execution further impeded progress.
    \item \textbf{Ineffective Marketing and Adoption Barriers:} A weak marketing strategy contributed to insufficient product validation and low market penetration.
\end{enumerate}
}
\textbf{Conclusion:} \\
\texttt{Founder C’s failure is largely due to misaligned product-market fit, regulatory hurdles, funding limitations, and execution deficiencies.}

\vspace{1em}
\paragraph{Founder D - Failure}  
\textbf{Raw Reasoning:} \\
\texttt{
Below is the reasoning for another failure case:
\begin{enumerate}
    \item \textbf{Lack of Relevant Experience:} The founder lacks a documented professional background or prior entrepreneurial ventures, severely limiting credibility.
    \item \textbf{Insufficient Domain Expertise:} Inadequate understanding of the target market led to an unsuitable product offering.
    \item \textbf{Limited Professional Network:} The absence of a robust network hampered access to funding and strategic partnerships.
    \item \textbf{Inadequate Product Validation and Adaptation:} The failure to validate and iterate the product concept resulted in persistent misalignment with market needs.
\end{enumerate}
}
\textbf{Conclusion:} \\
\texttt{The cumulative shortcomings in experience, domain expertise, and networking resulted in a LOW likelihood of success for Founder D.}

\subsubsection{After Refinement}

\paragraph{Final Refined Reasoning for Founder A}

Final Refined Reasoning:
\begin{enumerate}
    \item \textbf{Dual Expertise in Technology and Business:} The founder's academic background in both software engineering and business allowed for effective integration of technical insights with strategic market considerations.
    \item \textbf{Extensive Industry Experience and Operational Excellence:} Over 20 years in the tech industry provided invaluable hands-on experience and the ability to anticipate market trends.
    \item \textbf{Proven Entrepreneurial Track Record:} Successfully scaling and exiting a previous venture validated the founder's ability to overcome startup challenges.
    \item \textbf{Visionary Leadership and Strategic Insight:} Executive roles honed the founder's capacity to drive innovation and adapt operational strategies under high-pressure conditions.
    \item \textbf{Resilience through High-Pressure Decision Making:} Unique experiences, such as military service, further enhanced decisiveness and resilience.
    \item \textbf{Early Recognition of Technological Opportunities:} Leveraging emerging technologies to disrupt traditional industries was key in capturing market opportunities.
\end{enumerate}

\paragraph{Final Refined Reasoning for Founder B}

Final Refined Reasoning:
\begin{enumerate}
    \item \textbf{Solid Educational and Professional Foundation:} Advanced business education (MBA and bachelor's degree) underpinned strategic decision-making and financial discipline.
    \item \textbf{Deep Financial Services Expertise:} Broad experience across financial roles provided nuanced insights into market dynamics and risk management.
    \item \textbf{Effective Leadership and Communication:} Proficiency in communicating complex concepts built strategic relationships and investor trust.
    \item \textbf{Broad and Strategic Networking:} A well-developed network secured early adopter support and essential capital.
    \item \textbf{Market Alignment and Resilience:} Strategic positioning in the digital asset sector and an adaptable approach underpinned long-term success.
\end{enumerate}

\paragraph{Final Refined Reasoning for Founder C}

Final Refined Reasoning for Founder C:
\begin{enumerate}
    \item \textbf{Regulatory and Clinical Validation Barriers:} The product faced extensive regulatory requirements that delayed validation and market entry.
    \item \textbf{Market Integration Challenges:} Resistance from entrenched clinical practices led to insufficient market adoption.
    \item \textbf{Competitive and Capital Constraints:} Dominance by well-funded incumbents, combined with limited funding, hindered competitive positioning.
    \item \textbf{Misaligned Operational Execution:} The founder’s strategy, stemming from a corporate background, did not align with the agile execution needed in a startup.
\end{enumerate}

\paragraph{Final Refined Reasoning for Founder D}

Final Refined Reasoning for Founder D:
\begin{enumerate}
    \item \textbf{Insufficient Professional and Domain Credentials:} Lacking a robust background in the relevant field severely limited the founder’s strategic capabilities.
    \item \textbf{Weak Network and Resource Access:} A deficient professional network impeded access to critical funding and strategic partnerships.
    \item \textbf{Failure in Product Validation and Adaptation:} The inability to iterate and improve the product offering led to a persistent misalignment with market demands.
\end{enumerate}

\subsection{Extracted Decision Rules}

Based on the refined reasoning logs, the following decision rules were extracted:

\subsubsection{Raw Success Rules}
\begin{itemize}
    \item \textbf{Rule for Founder A:}\\
    \texttt{IF founder has a dual background in technology and business AND has over 20 years of industry experience with demonstrated operational excellence AND has proven entrepreneurial success (e.g., scaling a company to a significant exit) AND exhibits visionary leadership and strategic insight—including high-pressure decision-making skills, which may be enhanced by unique experiences such as military service) AND actively recognizes and leverages technological opportunities to innovate traditional industries THEN likelihood\_of\_success = HIGH.}
    \item \textbf{Rule for Founder B:}\\
    \texttt{IF founder holds both an MBA and a bachelor's degree in business AND has extensive experience in financial services—including roles ranging from Equity Research Associate to executive leadership (e.g., at organizations like Intercontinental Exchange) AND has demonstrated strong leadership and communication skills in investor relations and corporate communications AND has built a broad professional network with significant cross-industry exposure AND has effectively aligned traditional financial services strategies with innovative digital asset technologies AND has shown resilience and adaptability in high-pressure environments THEN likelihood\_of\_success = HIGH.}
\end{itemize}

\subsubsection{Raw Failure Rules}
\begin{itemize}
    \item \textbf{Rule for Founder C:}\\
    \texttt{IF founder's technology operates in the highly regulated medical devices field that requires both rigorous clinical evidence and lengthy regulatory clearance AND hospitals and surgical teams demonstrate resistance to integrating innovative solutions into their established workflows AND the competitive landscape is dominated by well-resourced incumbents facing capital and strategic partnership limitations AND the founder’s operational strategy—shaped by experience at large, structured organizations—is misaligned with the agile, risk-tolerant demands of a startup THEN likelihood\_of\_success = LOW.}
    \item \textbf{Rule for Founder D:}\\
    \texttt{IF founder has no documented professional experience AND lacks relevant domain-specific expertise AND has a limited professional network AND fails to validate product concepts effectively THEN likelihood\_of\_success = LOW.}
\end{itemize}

\subsection{Refined Extracted Decision Rules}
\textbf{Refined Rules for Success and Failure:}

\begin{enumerate}[label=\textbf{Rule \arabic*.}]
    \item \texttt{IF founder has a dual background in technology and business AND has over 20 years of industry experience with demonstrated operational excellence AND has proven entrepreneurial success (e.g., scaling a company to a significant exit) AND exhibits visionary leadership and strategic insight—including high-pressure decision-making skills, which may be enhanced by unique experiences such as military service)—AND actively recognizes and leverages technological opportunities to innovate traditional industries THEN likelihood\_of\_success = HIGH.}

    \item \texttt{"Refined rule:
    IF founder holds both an MBA and a bachelor's degree in business
      AND has extensive experience in financial services—including roles ranging from Equity Research Associate to executive leadership (e.g., at organizations like Intercontinental Exchange)
      AND has demonstrated strong leadership and communication skills in investor relations and corporate communications
      AND has built a broad professional network with significant cross‐industry exposure
      AND has effectively aligned traditional financial services strategies with innovative digital asset technologies
      AND has shown resilience and adaptability in high‐pressure environments
    THEN likelihood\_of\_success = HIGH"}

    \item \texttt{"Here’s the refined version of the rule with each condition clearly stated: IF founder’s technology operates in the highly regulated medical devices field that requires both rigorous clinical evidence and lengthy regulatory clearance AND hospitals and surgical teams demonstrate resistance to integrating innovative solutions into their established workflows AND the competitive landscape is dominated by well-resourced incumbents facing capital and strategic partnership limitations AND the founder’s operational strategy—shaped by experience at large, structured organizations—is misaligned with the agile, risk-tolerant demands of a startup THEN likelihood\_of\_success = LOW."}

    \item \texttt{"IF the startup's medical device operates in an environment with stringent regulatory and clinical validation requirements, AND entrenched clinical practices (e.g., resistance from hospitals and surgical teams) impede market adoption and integration, AND the competitive landscape is dominated by well-funded, established incumbents, AND the startup exhibits limitations in commercialization, partnership development, and fundraising capabilities,
    THEN likelihood\_of\_success = LOW."}
\end{enumerate}

\subsection{Decision Policy}

\subsubsection{Success Decision Policy}
Based on the extracted rules from successful founder profiles, the comprehensive decision policy is as follows:

\texttt{IF a founder demonstrates robust, cross-disciplinary academic excellence (e.g., advanced degrees in STEM, business, law, or medicine) AND shows proven domain expertise through hands-on experience in early R\&D, innovation, and product development that leads to measurable impact AND exhibits visionary leadership by effectively managing teams, scaling operations, and making strategic decisions that align technological innovation with clear market needs AND identifies and addresses critical market gaps with innovative, scalable solutions AND leverages a deep, diverse, and active professional network to secure strategic partnerships, mentorship, and capital AND integrates complementary skills (including financial, regulatory, and operational acumen) to navigate complex industry challenges across both emerging and traditional sectors THEN likelihood\_of\_success = HIGH.}

\subsubsection{Failure Decision Policy}
\textbf{Decision Policy for Predicting Startup Failure:}

\texttt{IF the founder’s background and expertise are misaligned with the startup’s specific domain requirements (e.g., coming from non-agile, overly structured, or non-technical roles) OR the startup pursues an overly ambitious or unfocused product vision that dilutes its core value proposition AND faces significant execution challenges – including inability to integrate or scale complex technologies in regulated or legacy environments, resource constraints, weak operational planning, and a lack of complementary, domain-specific leadership and strategic partnerships – AND the overall business strategy is misaligned with the founder’s core skills relative to the startup’s need for rapid, technical innovation and agile market adaptation THEN likelihood\_of\_success = LOW.}

\subsection{Refined Decision Policy}

\subsubsection{Success Decision Policy}
Based on the extracted rules from successful founder profiles, the comprehensive decision policy is as follows:
\texttt{IF a founder demonstrates robust, cross disciplinary academic excellence (e.g., advanced degrees in STEM, business, law, or medicine) AND shows proven domain expertise through hands on experience in early R\&D, innovation, and product development that leads to measurable impact AND exhibits visionary leadership by effectively managing teams, scaling operations, and making strategic decisions that align technological innovation with clear market needs AND identifies and addresses critical market gaps with innovative, scalable solutions AND leverages a deep, diverse, and active professional network to secure strategic partnerships, mentorship, and capital AND integrates complementary skills (including financial, regulatory, and operational acumen) to navigate complex industry challenges across both emerging and traditional sectors THEN likelihood\_of\_success = HIGH}

\subsubsection{Failure Decision Policy}
\textbf{Decision Policy for Predicting Startup Failure:}
\texttt{IF the founder's background and expertise are misaligned with the startup's specific domain requirements (e.g., coming from non-agile, overly structured, or non-technical roles) OR the startup pursues an overly ambitious or unfocused product vision that dilutes its core value proposition AND faces significant execution challenges including inability to integrate or scale complex technologies in regulated or legacy environments, resource constraints, weak operational planning, and a lack of complementary, domain specific leadership and strategic partnerships and the overall business strategy is misaligned with the founder's core skills relative to the startup's need for rapid, technical innovation and agile market adaptation THEN likelihood\_of\_success = LOW
}

\subsection{Refined Decision Policy after Simulated RL Scoring:}

Below is the final, detailed decision policy, refined with simulated RL feedback to minimize false positives:

\texttt{\textbf{\underline{RULESET: Likelihood\_of\_success = HIGH}}}

\texttt{A startup is predicted to have a HIGH likelihood of success only if ALL of the following rigorous criteria are met:}

\begin{enumerate}

\item\texttt{ \textbf{Academic \& Domain Expertise:} The founder demonstrates cross-disciplinary academic excellence (e.g., advanced degrees or specialized certifications) with proven technical, clinical, or operational contributions.}

\item \texttt{\textbf{Demonstrated Leadership \& Execution:} The founder exhibits visionary, adaptable leadership with a track record of managing and scaling operations, aligning technical insights with market needs.}

\item \texttt{\textbf{Market Alignment \& Strategic Focus:} The startup addresses a critical market gap with a focused, validated product vision and a well-defined commercial strategy.}

\item \texttt{\textbf{Complementary Operational \& Financial Acumen:} The founder or team integrates essential skills (financial, regulatory, operational) and is supported by a robust network for mentorship and capital.}

\end{enumerate}

\texttt{\textbf{\underline{RULESET: Likelihood\_of\_success = LOW}}}

\texttt{In any of the following scenarios, the prediction defaults to LOW:}

\begin{enumerate}

\item \texttt{\textbf{Misalignment of Expertise:} The founder’s background does not align with the startup’s domain needs.}

\item \texttt{\textbf{Overly Ambitious or Unfocused Vision:} The product vision is diluted or overly complex without a clear validated market fit.}

\item \texttt{\textbf{Execution \& Operational Challenges:} Significant difficulties exist in scaling technology, managing resources, or executing efficiently.}

\item \texttt{\textbf{Strategic Misalignment:} There is a disconnect between the founder’s core strengths and the operational demands of the startup.}

\end{enumerate}

\texttt{\textbf{Implementation \& Continuous Refinement:} This policy is continuously refined using empirical feedback and simulation results. Ongoing adjustments to the weights and thresholds ensure high precision, minimizing false positives while aligning with observed performance data. Initial test performance (e.g., an average RL Score near 0.051 and modest accuracy of 56.7\% noted from early sample predictions) confirms that minimal, ambiguous signals must be interpreted as LOW to reduce false positives. Predictions of HIGH success are reserved for those cases where all extracted signals robustly align. Ongoing empirical feedback and first test results will refine the weighting of each criterion. The decision thresholds and signals will be adjusted iteratively to ensure high precision and to further minimize any false positive predictions.}

\texttt{\textbf{Summary}}

\texttt{Only when a founder's advanced academic and domain expertise, visionary leadership backed by quantifiable hands-on experience, laser-focused product and market alignment, and an integrated operational and financial skill set are all demonstrably present and congruent with the startup's market and regulatory demands, is the prediction set to HIGH. In any case where there is misalignment, diluted focus, execution challenges, or resource and network deficiencies, the conservative outcome is LOW.}

\texttt{This comprehensive approach aims to ensure that only startups with a clear, validated, and strategically robust foundation are deemed likely to succeed, thereby minimizing false positives and aligning with observed test performance data.}

\clearpage

\end{document}